\pdfoutput=1

\documentclass[11pt]{article}
\usepackage[final]{acl}

\usepackage{amsmath,amsfonts,bm}

\def\eqref#1{equation~\ref{#1}}

\def\1{\bm{1}}

\DeclareMathAlphabet{\mathsfit}{\encodingdefault}{\sfdefault}{m}{sl}
\SetMathAlphabet{\mathsfit}{bold}{\encodingdefault}{\sfdefault}{bx}{n}

\usepackage{times}
\usepackage{latexsym}
\usepackage[T1]{fontenc}
\usepackage[utf8]{inputenc}

\usepackage{microtype}
\usepackage{inconsolata}
\usepackage{graphicx}
\usepackage{hyperref} 
\usepackage{booktabs}
\usepackage{float}
\usepackage{subcaption}
\usepackage{multirow}
\usepackage{makecell}
\usepackage{wrapfig}
\usepackage{amsmath}
\usepackage{amssymb}
\usepackage{xcolor}
\usepackage{xspace}
\usepackage{tikz}
\usepackage{enumitem,amssymb}
\usepackage{soul}
\usepackage{duckuments}
\usepackage[export]{adjustbox}
\usepackage{setspace}

\setlist[itemize]{leftmargin=5.5mm}

\usepackage[most]{tcolorbox}
\usepackage[frozencache,cachedir=minted-cache]{minted2}
\usepackage[framemethod=tikz]{mdframed}
\usepackage{listings}
\definecolor{mybrown}{RGB}{128,64,0}
\definecolor{red}{RGB}{235, 37, 64}
\definecolor{blue}{RGB}{64, 123, 211}

\NewDocumentCommand{\PromptTextBox}
  {
   O{mybrown!5!white}   
   O{mybrown!75!black}  
   m}          
{%
  \begin{minipage}[t]{\linewidth} \vspace{0pt}
    \centering
    \begin{tcolorbox}[colback=#1,colframe=#2,boxrule=1pt,
                      left=1pt,right=1pt,top=1pt,bottom=1pt,enhanced]
      \vspace*{0pt}
      \begin{spacing}{1.25}
      #3
      \end{spacing}\vspace*{0pt}
    \end{tcolorbox}
  \end{minipage}
}

\newcommand{\codeurl}{\url{https://github.com/ml-postech/Faithful-Agentic-XAI}}

\usepackage{setspace}
\usepackage{titlesec}
\titlespacing*{\paragraph}{0pt}{0.3em}{1em}
\setlist[itemize]{itemsep=0pt, leftmargin=5.5mm}

\title{Towards Faithful Agentic XAI: A Verification Method and an Open-World Benchmark for Better Model Faithfulness}

\author{
 \textbf{Jaechang Kim\textsuperscript{1}},
 \textbf{Sunung Mun\textsuperscript{1}},
 \textbf{Seungjoon Lee\textsuperscript{1}}, \\
 \textbf{Jaewoong Cho\textsuperscript{2}},
 \textbf{Jungseul Ok\textsuperscript{1}}
\\
\\
 \textsuperscript{1}POSTECH, 
 \textsuperscript{2}KRAFTON
}

\begin{document}
\maketitle

\begin{abstract}

Explainable AI (XAI) helps users interpret model behavior and identify potential faults.
Agentic XAI systems use Large Language Models (LLMs) to make explanations more accessible through natural-language interaction, but they can also produce plausible yet unfaithful explanations.
This risk arises because unreliable XAI outputs for complex models can be amplified by LLMs and mislead users.
We propose Faithful Agentic XAI (FAX), a verification method that improves the faithfulness of Agentic XAI explanations.
FAX decomposes draft explanations into claims and tests them with inherently faithful tools that query the target model directly, filtering unsupported or contradicted claims before final generation.
We also introduce CRAFTER-XAI-Bench, an open-world reinforcement learning benchmark with complex policies, diverse goals, and challenging scenarios for assessing model-specific faithfulness.
On CRAFTER-XAI-Bench, FAX improves simulation faithfulness from 0.20 for the strongest baseline to 0.46 while maintaining high informativeness, relevance, and fluency.
On three tabular benchmarks, FAX performs competitively with prior Agentic XAI baselines, but our analysis shows that these settings can conflate task accuracy with model-specific faithfulness.
These findings show that explicit verification is essential for faithful Agentic XAI and that faithfulness benchmarks must be designed to test explanations against the behavior of the target model itself.

\end{abstract}

\section{Introduction}

\begin{figure}
    \includegraphics[width=\linewidth]{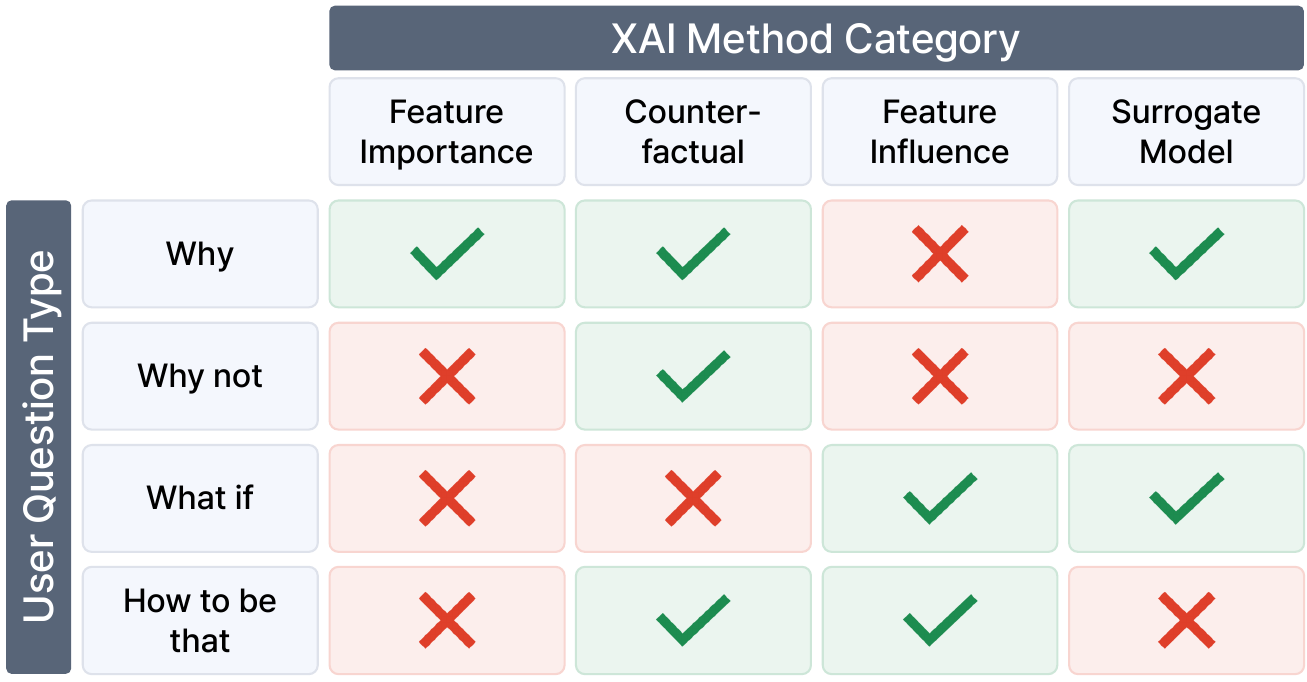}
\vspace{-1mm}
    \caption{
        Different XAI methods provide different information. 
        User question types are adopted from XAIQuestionBank~\citep{liao2020questionbank}.
    }
    \label{fig:xai_information_mapping}
\vspace{-3mm}
\end{figure}

Explainable AI (XAI) helps users interpret model behavior, diagnose potential faults, and decide when model outputs should be trusted.
Different XAI methods provide different information about model decisions, as shown in Figure~\ref{fig:xai_information_mapping}.
However, selecting an appropriate method and interpreting its output often requires expert-level knowledge of machine learning and XAI, creating a significant barrier for non-expert users.
To lower this barrier, Agentic XAI systems use a Large Language Model (LLM) to select suitable XAI methods and translate their outputs into natural-language explanations~\citep{slack23talktomodel, he2025conversational}.

This accessibility benefit introduces a critical risk: a fluent explanation is not necessarily a faithful one.
Existing Agentic XAI systems often assume that the selected XAI tools provide reliable evidence that can be safely verbalized.
While this assumption may be less visible in simple tabular settings, it breaks down for complex models and dynamic environments, where the unfaithfulness of XAI methods is a known issue~\citep{adebayo2018sanity}.
An LLM agent that naively trusts and rephrases unreliable tool outputs can turn noisy or spurious evidence into a coherent, plausible, and fundamentally incorrect explanation.
The central danger is therefore not merely that an agent may use imperfect tools, but that it can make unsupported claims sound authoritative to the user.
Whether an explanation is faithful cannot be judged from how plausible it sounds; it has to be checked against the behavior of the model being explained.

\begin{figure}
    \includegraphics[width=\linewidth]{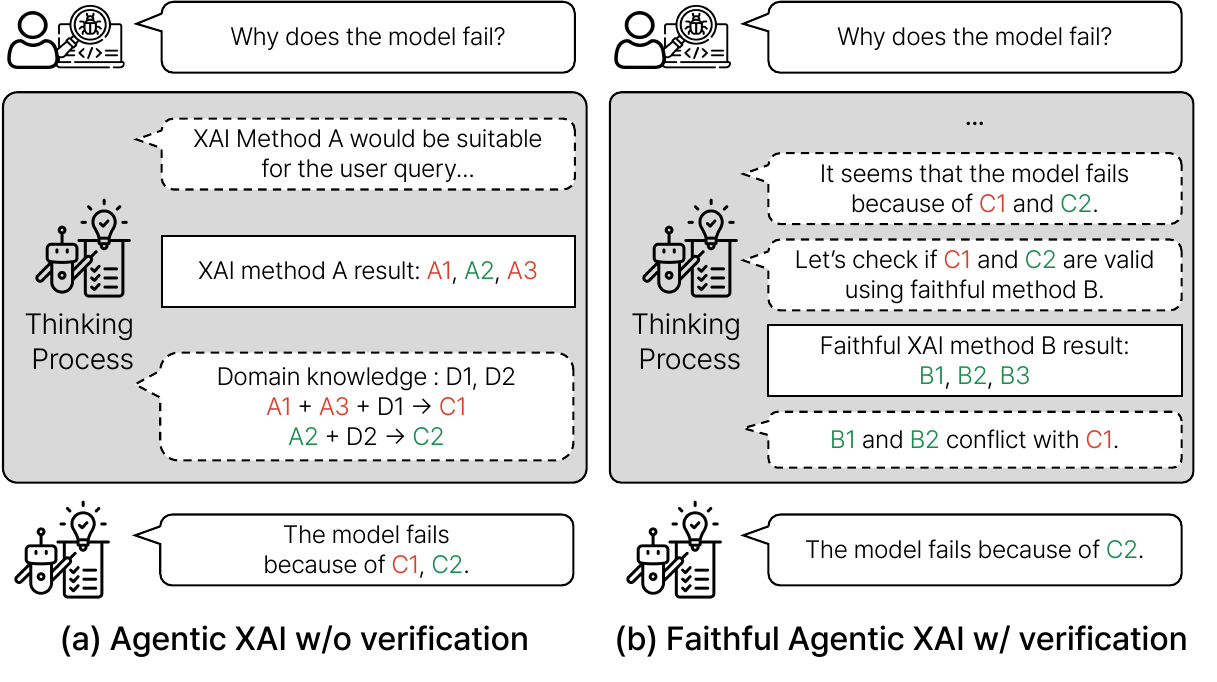}
    \caption{
        (a) Agentic XAI uses XAI methods suited to a user query and generates a natural-language response.
        (b) FAX verifies response claims with inherently faithful XAI methods.
    }
    \label{fig:motivation}
\end{figure}

We address this gap by proposing Faithful Agentic XAI (FAX), a verification method for Agentic XAI that treats a drafted explanation not as an answer but as a set of hypotheses about the target model.
FAX decomposes the draft into testable claims, analyzes the evidence supporting each claim, and tests uncertain claims with inherently faithful tools, whose evidence is computed by directly querying the target model under a given input, intervention, or counterfactual condition.
Each claim thus makes a prediction that the model can contradict: if the draft insists that a feature drives the decision, editing that feature should change the decision.
The agent therefore does more than verbalize tool outputs; it interrogates the model it explains, and the test results control final generation: corroborated claims are retained, refuted claims are removed or rewritten, and inconclusive claims are not presented as verified facts.
Figure~\ref{fig:motivation} illustrates this motivation and our approach.

Evaluating faithful Agentic XAI also requires benchmarks that separate model-specific faithfulness from generic task knowledge.
Standard tabular benchmarks remain useful for comparison with prior conversational XAI systems, but they can obscure the distinction between explaining the target model and predicting the task label.
Because these datasets are familiar binary classification tasks, an LLM may achieve a high simulation score by relying on dataset-level priors or task accuracy rather than on a faithful explanation of the learned model.
We therefore introduce CRAFTER-XAI-Bench, a scalable benchmark built on an open-world reinforcement learning (RL) environment.
The benchmark contains complex policies with diverse goals and challenging scenarios where generic environment knowledge is insufficient; a faithful explanation must capture the behavior of the specific target policy.
CRAFTER-XAI-Bench evaluates explanations by the same principle that FAX uses to build them: it derives behavioral predictions from an explanation and checks them against the target policy's actual decisions, together with usability-oriented metrics for informativeness, query relevance, and fluency.

To summarize our main contributions:
\begin{itemize}
\item We propose FAX, a verification method for Agentic XAI that decomposes draft explanations into claims and tests them against the target model with inherently faithful tools before they reach the user.
\item We introduce CRAFTER-XAI-Bench, an open-world RL benchmark that evaluates model-specific explanation faithfulness by checking the behavior an explanation predicts against the target policy under diverse and challenging user queries.
\item We evaluate FAX on CRAFTER-XAI-Bench and three tabular benchmarks following TalkToModel, showing that FAX improves faithfulness in the primary open-world setting, that this gain comes from claim verification rather than from rewriting alone, and that familiar tabular evaluations can conflate model faithfulness with task accuracy.
\end{itemize}

\section{Related Work}

\paragraph{XAI tools and agentic workflows}
Post-hoc XAI methods provide complementary but imperfect evidence, including attributions and saliency~\citep{simonyan2014saliency}, surrogate rules~\citep{ribeiro2018anchors, ribeiro2016lime}, examples or counterfactuals~\citep{chen2019protopnet, wachter2018counterfactual}, and concepts~\citep{TCAV, yuksekgonul23posthoc}.
For agentic XAI, the key distinction is whether a tool directly queries the target model under a specified input, intervention, or counterfactual condition, or instead provides an approximate proxy that may fail sanity checks~\citep{adebayo2018sanity}.
Dashboards and conversational XAI systems lower the expertise barrier by aggregating tools or mapping user questions to XAI operations~\citep{explainerdashboard, wenzhuo22omnixai, aix360, Zytek24explingo, castelnovo2024augmenting, Zhang25followup, slack23talktomodel, he2025conversational}.
Taking interactivity further, \citet{kim2025agentic} advocate agentic interpretability, in which an LLM proactively helps users understand it through multi-turn conversation.
Agentic XAI, where the LLM explains another model, shares this concern: existing systems largely verbalize selected tool outputs as reliable evidence.
FAX addresses this gap by treating draft explanation claims as hypotheses and verifying them with inherently faithful tools before final generation.

\paragraph{Faithfulness evaluation}
Faithfulness is commonly evaluated through \emph{simulatability}: whether an explanation helps an observer reproduce or predict the target model's behavior~\citep{jacovi2020towards, lyu2024towards}.
Prior work uses student models~\citep{li2020evaluating} or human simulators~\citep{chen2018learning, nguyen2018comparing, hase2020evaluating}.
We follow the same principle with an LLM simulator, but judge its predictions against actual target-model outputs on unseen states.
Appendix~\ref{appendix:extended_related_work} extends this discussion.

\begin{figure}[t]
    \centering
    \includegraphics[width=\linewidth]{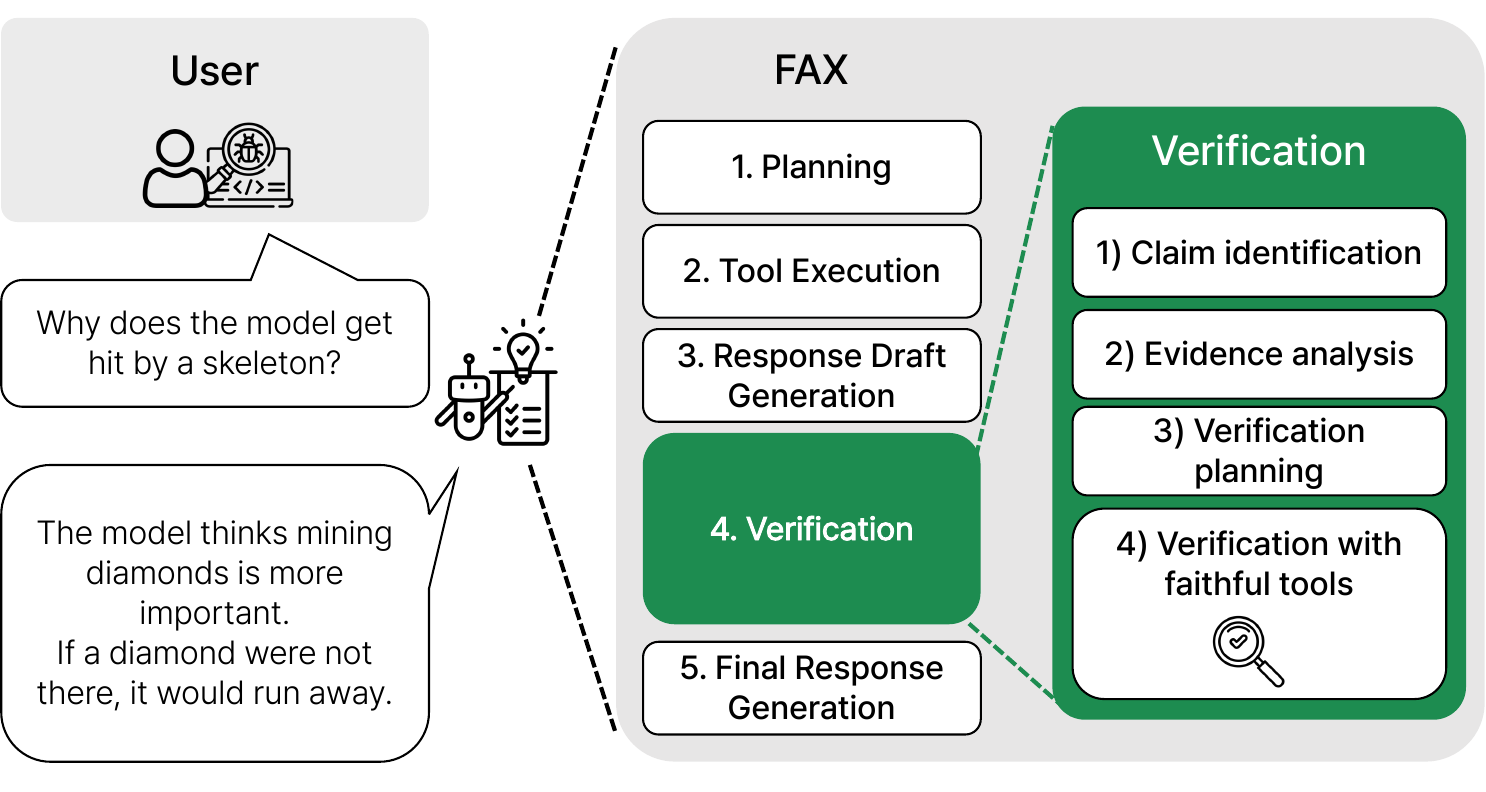}
    \caption{FAX augments an agentic XAI pipeline with an explicit verification step that detects and removes unfaithful claims before final generation.}
    \label{fig:method_overview}
\end{figure}

\section{Method: Faithful Agentic XAI}
\label{sec:method}

\begin{figure*}[t]
    \centering
    \includegraphics[width=0.95\linewidth]{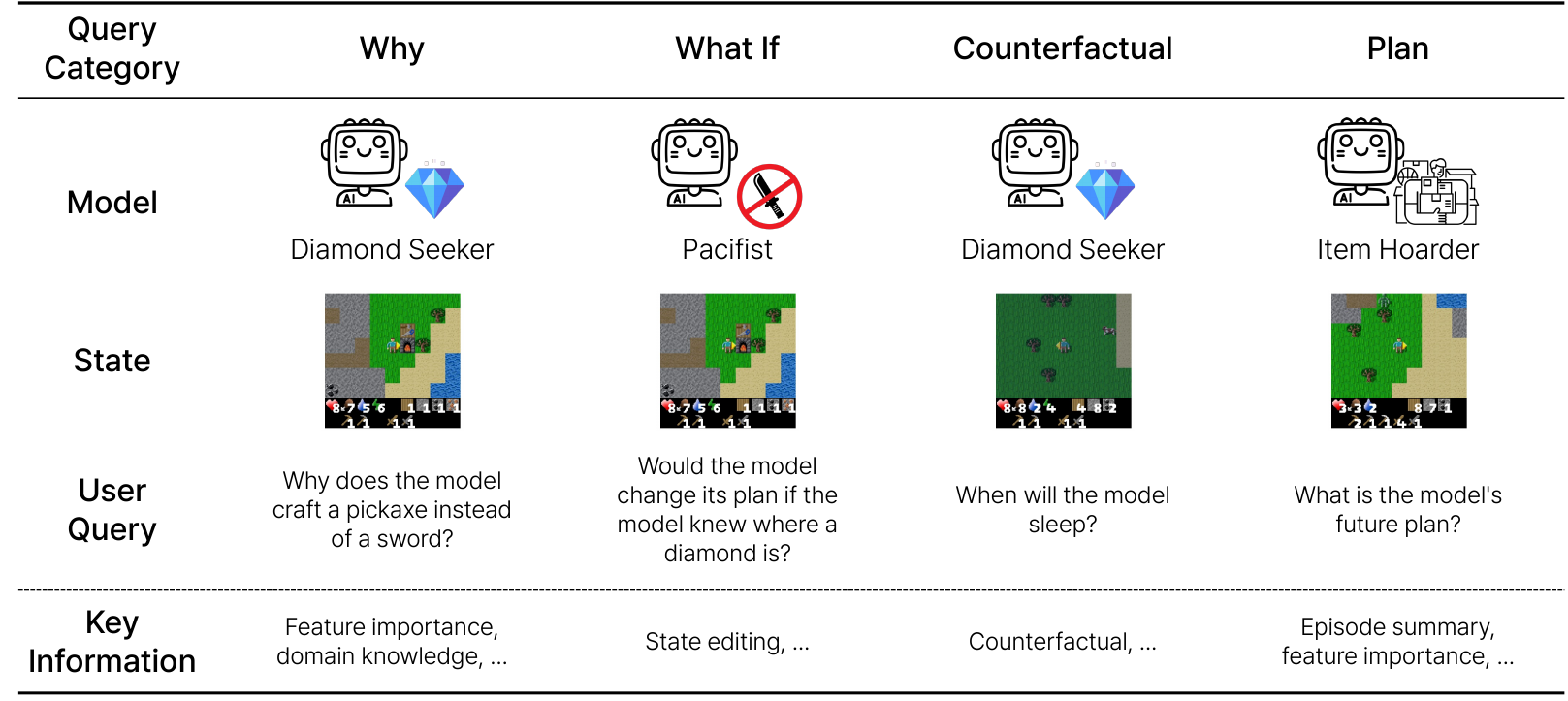}
    \caption{
        Evaluation scenarios are grouped into four categories.
        Each category represents a different kind of query, and different information is useful for answering each query type.
    }
    \label{fig:eval_scenario}
    \vspace{-3mm}
\end{figure*}

We propose \textbf{Faithful Agentic XAI (FAX)}, a verification method that improves the faithfulness of agentic natural-language explanations of complex models.
FAX keeps the usual agentic pipeline of planning, tool use, and drafting, and inserts a verification step that audits draft claims before final generation.
The draft is treated as a set of hypotheses about the target model, and the agent, rather than the user, bears the burden of testing whether conclusions drawn from noisy XAI outputs hold for that model.

\subsection{Agentic XAI pipeline}
Our methodology builds on Agentic XAI, where an LLM controller orchestrates XAI tools \cite{slack23talktomodel, he2025conversational}.
Given a user query $q$, model policy $\pi$, and state $s_t$, the agent generates an explanation $E$ for the model decision $a_t = \pi(s_t)$.
Figure~\ref{fig:method_overview} illustrates the overall FAX workflow.

\paragraph{Overall workflow}
FAX has five stages.
\textit{Planning} identifies the needed information; \textit{Tool Execution} calls selected XAI tools and converts outputs to text; \textit{Response Drafting} produces an initial explanation $E_{draft}$; \textit{Verification using Faithful Tools} labels draft claims as supported, contradicted, or unresolved; and \textit{Final Response Generation} produces $E_{final}$ from the draft and verification report.

\paragraph{Why the draft must be verified}
The draft $E_{draft}$ is useful as a concrete explanation candidate, but it can inherit noisy correlations from post-hoc tools such as SHAP or introduce plausible but incorrect causal links from the LLM.
FAX therefore treats the claims in $E_{draft}$ as hypotheses to check, rather than explanations to deliver directly.
Post-hoc tools and the LLM can propose such claims but cannot confirm them; a meaningful check must be able to contradict a claim, which requires querying the target model itself.

\subsection{Faithful verification}
\label{subsec:verification}

The verification step takes $E_{draft}$ as input and returns a structured audit report for final generation.
Unlike generic self-reflection, which asks the LLM whether its own explanation sounds plausible and does not improve faithfulness, FAX forces the agent to identify concrete claims, inspect where their support came from, and test uncertain claims with tools whose outputs are directly tied to the target model.

\paragraph{Step 1: Claim identification}
The agent parses $E_{draft}$ into testable atomic claims $\mathcal{C} = \{c_1, c_2, \dots, c_n\}$ about the model's behavior.
Each claim should be specific enough to be checked against evidence, such as a statement that a state feature, resource, or event affects the model action.
For example, a Crafter draft may claim: ``\textit{The agent chose to make a stone pickaxe because it prioritized gathering iron.}''

\paragraph{Step 2: Supporting evidence analysis}
For each claim $c_i$, the agent records the evidence used to support it and separates evidence sources by reliability.
A claim may already be grounded by faithful-tool evidence, may be suggested only by noisy post-hoc tools ($\mathcal{T}_{noisy}$), or may come from the LLM's prior knowledge while verbalizing tool outputs.
The agent also checks context consistency, such as whether a claimed missing resource is actually present in the observed state.
Claims that depend on noisy evidence, LLM priors, or inconsistent context are marked as requiring verification.

\paragraph{Step 3: Verification planning}
For each claim requiring verification, the agent plans calls to inherently faithful tools ($\mathcal{T}_{faithful}$), such as State Editing and Counterfactuals, with up to three calls per claim.
The most informative tests are those whose outcome the claim predicts, i.e., an intervention or counterfactual condition under which the model behavior should change, or stay the same, if the claim is correct.
Following the logic of falsification~\citep{popper1959logic}, a claim earns credit only by surviving a test that could have contradicted it; when no such test is feasible, the agent gathers additional faithful evidence and leaves the claim uncertain unless that evidence is strong.

\begin{figure*}[t]
    \centering
    \includegraphics[width=0.8\linewidth]{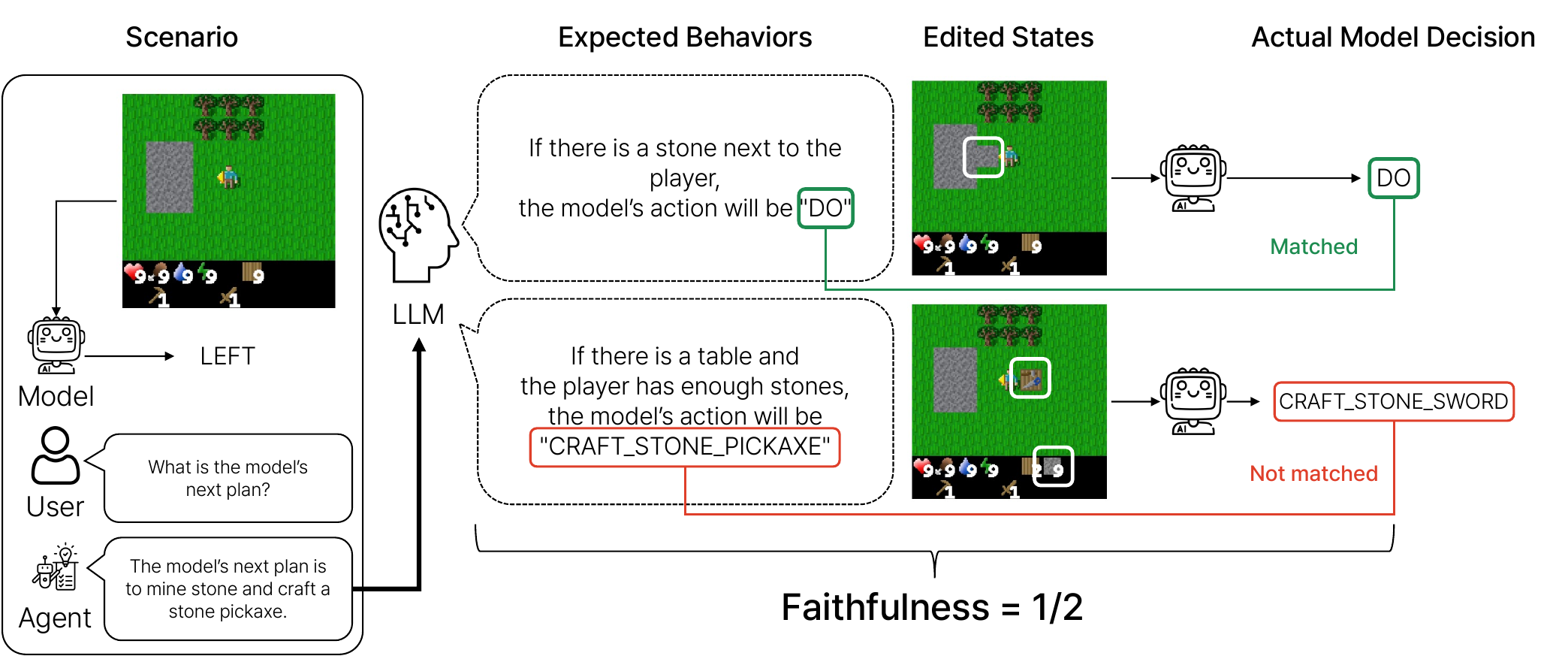}
    \caption{
        Faithfulness is evaluated by simulation accuracy.
        The LLM evaluator predicts the model decision for an unseen state based on the text explanation.
    }
    \label{fig:faithfulness_metric}
\end{figure*}

\paragraph{Inherently faithful tools}
A tool is inherently faithful when the core claim of its evidence can be judged true or false by directly executing the target model $\pi$ and the explanation uses that model-executed result rather than a surrogate inference.
State Editing returns the model decision $\pi(s'_t)$ on an edited state $s'_t$, and counterfactual tools accept candidate edits only if the target model's decision changes as specified.
Valid outputs of these tools are faithful by construction, although they may be less informative than richer post-hoc explanations. We provide detailed definition in  Appendix~\ref{appendix:inherently_faithful_tools}.

For the stone-pickaxe claim above, the agent may search for counterfactual changes that alter the action or call \texttt{edit\_state} to remove iron-related features and observe $\pi(s'_t)$.

\paragraph{Step 4: Verification with faithful tools}  
The agent executes the planned calls and observes $\pi(s'_t)$.
Each tested claim is assigned one of three statuses.
{Refuted} means faithful evidence contradicts the claim; for example, removing the allegedly critical feature leaves the action unchanged.
{Corroborated} means the faithful-tool result is consistent with the claim, such as an intervention changing behavior in the predicted direction.
{Inconclusive} means the test is invalid, fails, or remains ambiguous, so the claim should not be regarded as verified.

\paragraph{Using verification results}
The verification report records each claim, its original supporting evidence, the faithful-tool test, and the resulting status.
Final response generation uses this report as a control signal: corroborated claims can be retained and grounded, refuted claims are removed or rewritten, and inconclusive claims are omitted or expressed with uncertainty.
When noisy post-hoc evidence conflicts with faithful-tool evidence, the final response prioritizes the faithful-tool result.
Thus, verification directly determines which draft claims survive into the user-facing explanation.

\begin{table*}[t]

\caption{
    Results on CRAFTER-XAI-Bench (41 scenarios).
    FAX improves faithfulness while preserving informativeness, query relevance, and fluency compared to all baselines.
    Each cell is the mean over scenarios and three runs; $\pm$ denotes the sample standard deviation of the average across the three runs.
    Reflection-only is an ablation of FAX that keeps final synthesis but replaces claim verification with a generic reflection step.
    The best method in each metric is denoted with \textbf{boldface}.
}
\label{tab:automatic_faithfulness}
\centering

\resizebox{0.9\textwidth}{!}{%
\begin{tabular}{ccclcccc}
\toprule
\textbf{Method} & \textbf{\makecell{ Use \\ structured \\ workflow?}}  & \textbf{\makecell{Use \\verification \\ stage?}} & \textbf{Query Category} & \textbf{Faithfulness} & \textbf{Informativeness} & \textbf{Query Relevance} & \textbf{Fluency} \\
\midrule

\multirow{5}{*}{\makecell[c]{Explainer \\ Dashboard}} 
& \multirow{5}{*}{N/A} & \multirow{5}{*}{N/A} 
 & Counterfactual     & 0.14 & 0.27 & 0.31 & 0.26 \\
 &&& What if          & 0.19 & 0.25 & 0.36 & 0.26 \\
 &&& Plan             & 0.14 & 0.34 & 0.48 & 0.26 \\
 &&& Why              & 0.31 & 0.32 & 0.45 & 0.26 \\
 \cmidrule(lr){4-8}
 &&& \textbf{Average} & 0.20{\scriptsize$\pm$0.04} & 0.29{\scriptsize$\pm$0.00} & 0.40{\scriptsize$\pm$0.00} & 0.26{\scriptsize$\pm$0.00} \\
\midrule

\multirow{5}{*}{Naive LLM} 
& \multirow{5}{*}{\Large $ \times$} & \multirow{5}{*}{\Large $\times$} 
 & Counterfactual     & 0.11 & 0.77 & 0.95 & 0.99 \\
 &&& What if          & 0.17 & 0.91 & 0.98 & 0.99 \\
 &&& Plan             & 0.17 & 0.82 & 0.99 & 0.99 \\
 &&& Why              & 0.13 & 0.91 & 1.00 & 0.99 \\
 \cmidrule(lr){4-8}
 &&& \textbf{Average} & 0.14{\scriptsize$\pm$0.03} & 0.85{\scriptsize$\pm$0.02} & 0.98{\scriptsize$\pm$0.00} & \textbf{0.99}{\scriptsize$\pm$0.00} \\
\midrule

\multirow{5}{*}{\makecell[c]{ Unstructured \\ Agentic XAI }} 
& \multirow{5}{*}{\Large $\times$} & \multirow{5}{*}{$\triangle$} 
 & Counterfactual     & 0.12 & 0.91 & 0.98 & 0.99 \\
 &&& What if          & 0.34 & 0.90 & 0.99 & 0.98 \\
 &&& Plan             & 0.17 & 0.86 & 0.97 & 0.99 \\
 &&& Why              & 0.08 & 0.90 & 1.00 & 0.99 \\
 \cmidrule(lr){4-8}
 &&& \textbf{Average} & 0.18{\scriptsize$\pm$0.04} & 0.89{\scriptsize$\pm$0.02} & 0.98{\scriptsize$\pm$0.01} & \textbf{0.99}{\scriptsize$\pm$0.00} \\
\midrule

\multirow{5}{*}{\makecell[c]{Structured \\ Agentic XAI \\ w/o verification}} 
& \multirow{5}{*}{$\bigcirc$} & \multirow{5}{*}{\Large $\times$} 
 & Counterfactual     & 0.11 & 0.92 & 0.99 & 0.99 \\
 &&& What if          & 0.28 & 0.90 & 1.00 & 0.98 \\
 &&& Plan             & 0.15 & 0.86 & 0.99 & 0.99 \\
 &&& Why              & 0.13 & 0.91 & 1.00 & 0.99 \\
 \cmidrule(lr){4-8}
 &&& \textbf{Average} & 0.17{\scriptsize$\pm$0.01} & \textbf{0.90}{\scriptsize$\pm$0.01} & \textbf{0.99}{\scriptsize$\pm$0.00} & \textbf{0.99}{\scriptsize$\pm$0.00} \\
\midrule

\multirow{5}{*}{\makecell[c]{Reflection-only }} 
& \multirow{5}{*}{$\bigcirc$} & \multirow{5}{*}{\Large $\times$} 
 & Counterfactual     & 0.19 & 0.93 & 0.98 & 0.95 \\
 &&& What if          & 0.30 & 0.86 & 0.98 & 0.95 \\
 &&& Plan             & 0.11 & 0.74 & 0.99 & 0.98 \\
 &&& Why              & 0.11 & 0.83 & 0.98 & 0.98 \\
 \cmidrule(lr){4-8}
 &&& \textbf{Average} & 0.18{\scriptsize$\pm$0.03} & 0.84{\scriptsize$\pm$0.01} & 0.98{\scriptsize$\pm$0.00} & 0.96{\scriptsize$\pm$0.02} \\
\midrule

\multirow{5}{*}{\makecell[c]{FAX (proposed)}} 
& \multirow{5}{*}{$\bigcirc$} & \multirow{5}{*}{$\bigcirc$} 
 & Counterfactual     & 0.35 & 0.93 & 0.94 & 0.95 \\
 &&& What if          & 0.48 & 0.89 & 0.99 & 0.97 \\
 &&& Plan             & 0.48 & 0.86 & 0.99 & 0.98 \\
 &&& Why              & 0.54 & 0.92 & 0.99 & 0.98 \\
 \cmidrule(lr){4-8}
 &&& \textbf{Average} & \textbf{0.46}{\scriptsize$\pm$0.04} & \textbf{0.90}{\scriptsize$\pm$0.02} & 0.98{\scriptsize$\pm$0.01} & 0.97{\scriptsize$\pm$0.01} \\
\bottomrule
\end{tabular}
}
\end{table*}

\begin{table*}[t]
\centering
\caption{
    Evaluation on the tabular benchmarks used by TalkToModel, under the original model and a feature-ablated model variant in which the top prior-aligned feature is removed before retraining.
    The columns with feature-ablated model is the evaluation without the prior in the LLM.
    All metrics are rescaled to 0-1.
    Note that the expected faithfulness of a random explanation is 0.5.
    Full table is in Appendix~\ref{appendix:tabular_full}.
}
\label{tab:tabular_results}
\resizebox{0.9\textwidth}{!}{%
\begin{tabular}{ccccccccc}
\toprule
 &  \multicolumn{4}{c}{Original Model} & \multicolumn{4}{c}{Feature-Ablated Model} \\
\cmidrule(lr){2-5} \cmidrule(lr){6-9}
Method &  Faithful. & Informat. & Query Rel. &  Fluency  & Faithful.& Informat. & Query Rel. &  Fluency \\
\midrule

{\makecell{Explainer Dashboard}}
& 0.57 &   0.46 & 0.49 & 0.45
           & 0.67 &  0.43 & 0.48 & 0.54  \\

{TalkToModel}
  & 0.65 & 0.26 & 0.34 & 0.43
           & 0.65 & 0.21 & 0.28 & 0.45 \\

{Naive LLM}
 & \textbf{0.74} &\textbf{0.81}  & \textbf{0.89} &0.99 
           & 0.60 &\textbf{0.80} & 0.85 &0.99   \\

{\makecell[c]{ Unstructured  Agentic XAI }}
  & 0.68 &  0.58 & 0.68 & 0.86 & 0.67 & 0.56 & 0.69 & 0.89 \\

{\makecell[c]{Structured  Agentic XAI  w/o verification}}
 & 0.71 &  0.71 & 0.82 & 0.99 & 0.67 & 0.71 & 0.85 & 0.99 \\

{FAX (proposed)}
 & 0.70 &0.65 & 0.86 & \textbf{1.00}
           & \textbf{0.71} & 0.72 & \textbf{0.91} & \textbf{1.00} \\

\bottomrule
\end{tabular}
}
\end{table*}

\section{CRAFTER-XAI-Bench}

We introduce CRAFTER-XAI-Bench to evaluate agentic XAI across diverse and scalable scenarios without relying on human evaluation.

\subsection{Setting}

\paragraph{Environment}
We use Crafter~\citep{hafner2021crafter}, an open-world RL environment that requires long-term planning over high-dimensional states.
This setting supports policies with distinct behaviors and challenges XAI because actions can depend on long-horizon context.

\paragraph{XAI tools}
We use four representative tools: SHAP~\citep{shap} for feature attribution, MACE~\citep{karimi2020mace} for counterfactuals, HIGHLIGHTS~\citep{amir2018highlights} for episode saliency, and State Editing for interventions.
MACE and State Editing are treated as inherently faithful in our setting because their outputs are validated by direct target-model decisions, while SHAP and HIGHLIGHTS provide useful but potentially noisy post-hoc evidence.
In FAX, all four tools are available during tool execution, while the verification step can call only the inherently faithful ones.

\paragraph{Models}
We train three policies with different reward shaping: \textit{Diamond Seeker} emphasizes diamond-related achievements, \textit{Item Hoarder} rewards larger inventories, and \textit{Pacifist} penalizes attacking monsters.
High-quality explanations should reveal these policy-specific preferences rather than generic Crafter knowledge.

\subsection{Evaluation scenarios}

We evaluate four query categories: why, what-if, counterfactual, and plan.
Each scenario consists of a model, a state, and a user query; the benchmark contains 41 scenarios (10 why, 11 what-if, 10 counterfactual, and 10 plan).
Figure~\ref{fig:eval_scenario} shows examples, and the full scenario list is in Appendix~\ref{appendix:query_list}.

\subsection{Evaluation metrics}

We evaluate explanations with simulational model faithfulness and LLM-based usability metrics.

\paragraph{Simulational model faithfulness}
Simulation accuracy measures whether an explanation lets an evaluator predict the target model's behavior on unseen states (Figure~\ref{fig:faithfulness_metric}).
The faithfulness evaluator uses the generated explanation as a behavioral hypothesis about the target model.
Given the scenario state, the user question, and the explanation, the evaluator LLM identifies claims that imply how the model should act under related but unseen states and writes five hypotheses, each consisting of the claim, a state edit, and the action the explanation predicts after the edit (Figure~\ref{fig:appendix_system_prompt_faith}).
Each edit is applied to the scenario state with State Editing, the target policy is executed on the edited state, and a hypothesis is correct if the predicted action equals the executed action; the score of an explanation is the fraction of correct hypotheses.
The evaluator LLM proposes the tests and the predictions, but it does not judge whether they are correct: the reference labels come from executing the target model.
The metric therefore does not measure agreement with an LLM's opinion of the explanation, but whether the behavior that the explanation implies actually occurs.

\paragraph{Claim-dependent tests}
Because the hypotheses are derived from each explanation, different explanations are tested on different edited states.
This follows counterfactual simulatability~\citep{chen2024models, limpijankit2025counterfactual}, which evaluates an explanation on the counterfactual inputs that the explanation itself speaks to and distinguishes \emph{precision}, whether the explanation's implied predictions hold, from \emph{generality}, how much of the model's behavior the explanation covers.
Our faithfulness score corresponds to precision, and we report informativeness (below) as a complementary measure of how much decision-relevant content an explanation provides.
A consequence is that test difficulty can vary between explanations: an explanation that states a few simple, directly tested behaviors may induce easier hypotheses than one that describes a broader policy.
Faithfulness should therefore be read together with informativeness, and the absolute score (e.g., 0.46) is a simulation accuracy under this protocol, suited to comparing methods with the same evaluator, not the fraction of true statements in an explanation.

\paragraph{Sources of evaluator error}
The evaluator can still introduce errors when generated states are too difficult to reason about, when state modifications are invalid or underspecified, or when the judge model relies on its own priors.
For this reason, Appendix~\ref{appendix:additional_quantitative_analyses} reports a cross-model judge check with Qwen3-32B and Qwen3-8B.

\paragraph{LLM-based usability metrics}
We score informativeness, query relevance, and fluency as usability-oriented metrics.
Informativeness measures decision-relevant content, query relevance measures whether the response answers the user query, and fluency measures organization and grammaticality.
We adapt the G-Eval-style scalar interpretation protocol~\citep{geval} with task-specific rubrics; prompts are in Appendix~\ref{appendix:prompt_eval}.

\section{Experiments}

    \begin{figure}[t]
        \centering
        \includegraphics[width=\linewidth]{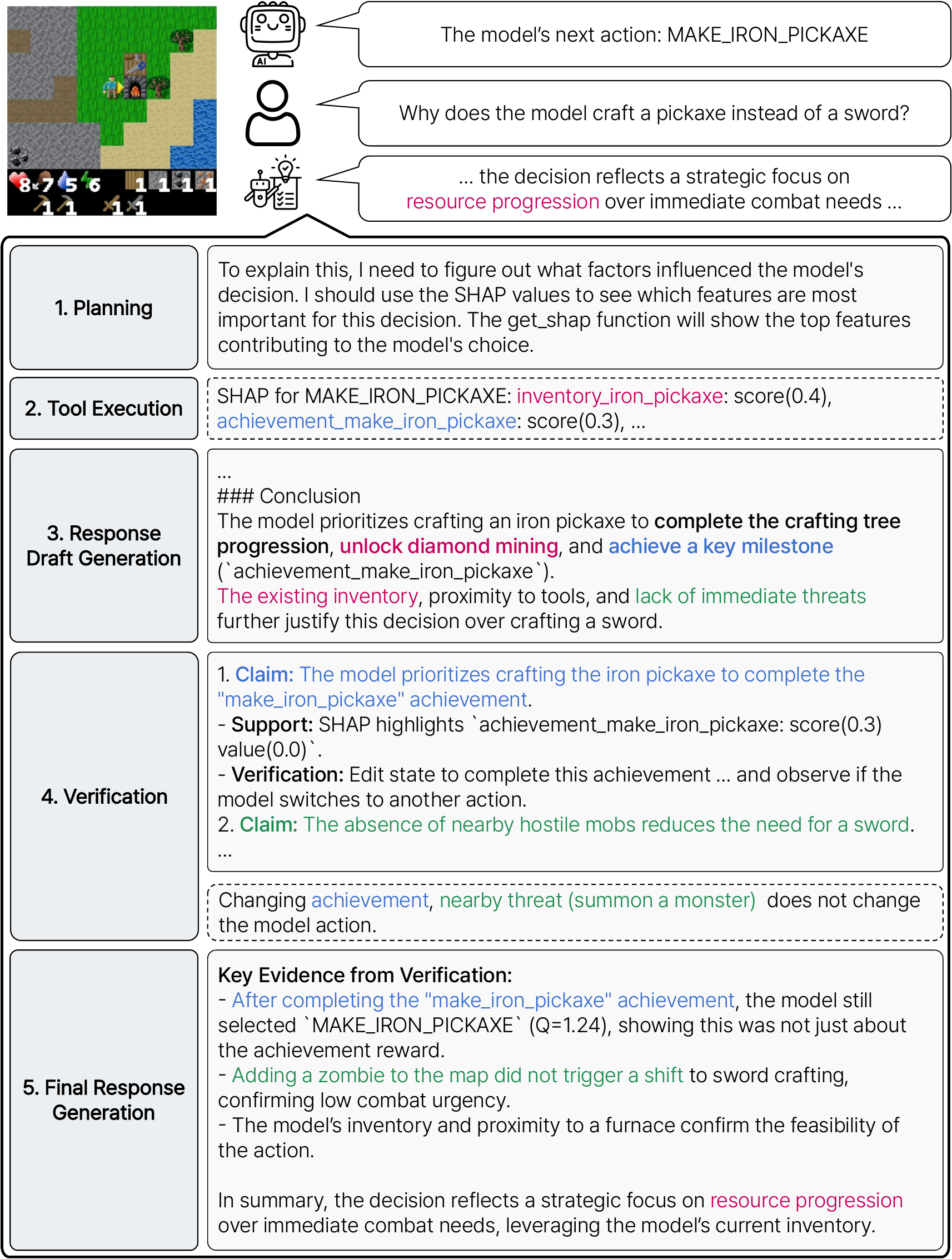}
        \caption{
            Each claim in the response draft is verified using faithful tools.
            Corresponding contents are color-coded, and unnecessary parts are replaced with ``...'' for readability.
        }
        \vspace{-3mm}
        \label{fig:qualitative_verification_ablation}
    \end{figure}

    \begin{figure*}[t]
        \centering
        \includegraphics[width=0.9\linewidth]{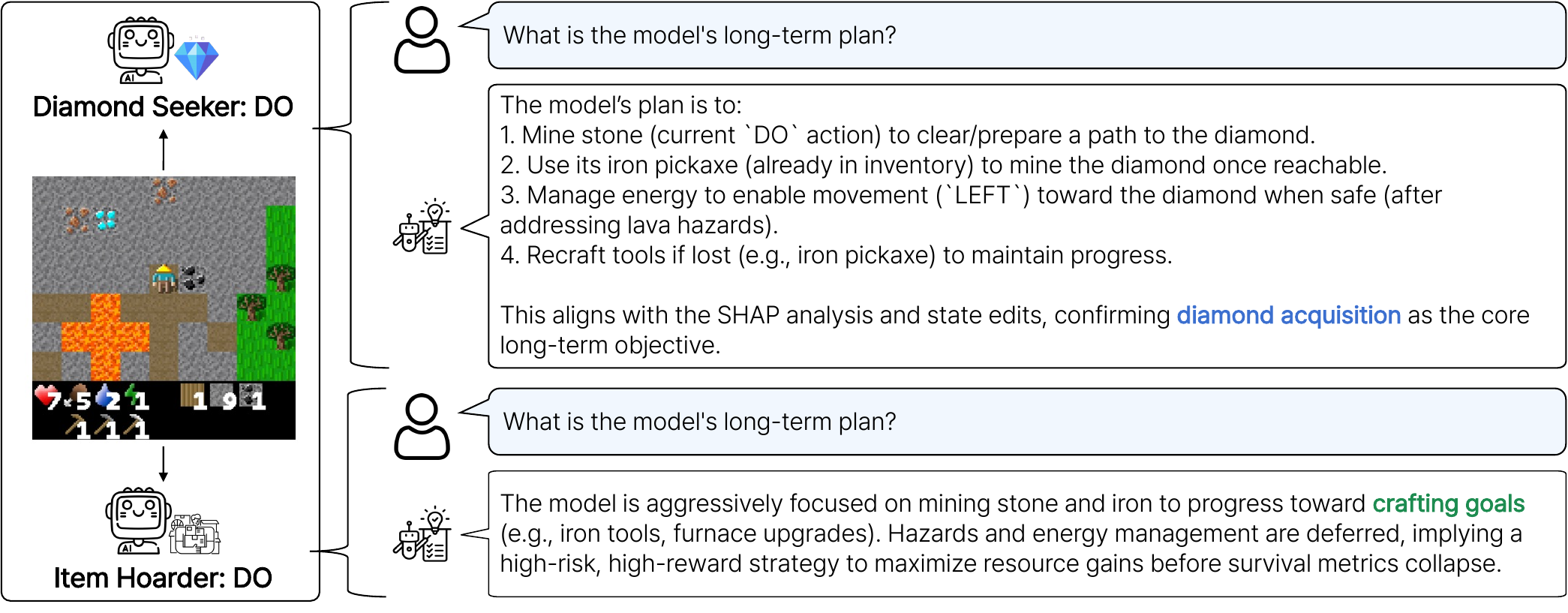}
        \caption{
            Different models produce different explanations even when the state and model action are the same.
            This example highlights why CRAFTER-XAI-Bench is necessary: the benchmark tests whether an explanation captures model-specific behavior rather than only plausible domain knowledge.
        }
        \label{fig:analysis_model_distinction}
    \end{figure*}

\subsection{Experimental setting}

\paragraph{Evaluation settings}
We evaluate FAX in two complementary settings.
The primary setting is CRAFTER-XAI-Bench, which tests model-specific faithfulness in an open-world RL environment where generic domain priors are insufficient for answering the user queries.
The second setting follows the tabular benchmarks used by TalkToModel~\citep{slack23talktomodel}, covering Pima Indians Diabetes and German Credit from the UCI Machine Learning Repository~\citep{dua2017uci} and the COMPAS recidivism dataset from ProPublica's Machine Bias investigation~\citep{angwin2016machinebias}.
This tabular setting connects our evaluation to prior conversational XAI work and tests whether FAX remains useful when the agent must reason over noisy post-hoc explanation evidence.
We explain TalkToModel's gradient-boosted classifiers with its XAI operations on 12 queries per dataset; faithfulness asks whether an explanation predicts the direction in which an edit moves the predicted probability, so chance level is 0.5 (Appendix~\ref{appendix:tabular_full}, which also notes that two baselines did not receive the dataset description).

We compare FAX against five baselines: ExplainerDashboard~\citep{explainerdashboard}, a non-agentic baseline that directly presents multiple XAI outputs and excludes State Editing in CRAFTER-XAI-Bench because it requires a concrete edit instruction; TalkToModel~\citep{slack23talktomodel}, a parser-based conversational XAI reference for tabular models; Naive LLM, which uses no XAI tools and relies on the backbone LLM and domain prompt; Unstructured Agentic XAI, a tool-using agent without a predefined workflow or mandatory verification, inspired by~\citep{he2025conversational}; and Structured Agentic XAI w/o Verification, a FAX ablation that keeps planning, tool execution, and drafting but omits verification and synthesis.
Full baseline descriptions are provided in Appendix~\ref{appendix:baseline_descriptions}.

\paragraph{Implementation details}
On CRAFTER-XAI-Bench, we use Qwen3-32B~\citep{qwen3} as the backbone LLM for FAX and the LLM-based baselines and as the evaluator.
On the tabular benchmarks, all LLM-based methods and the evaluator use Qwen3-8B.
The agentic workflows are implemented using LangGraph~\citep{langgraph}.
Prompts and further details are in Appendices~\ref{appendix:prompt_xai} and~\ref{appendix:implementation_details}.
On CRAFTER-XAI-Bench, each method answers every scenario in three independent runs.
Code, policy checkpoints, and outputs are available at \codeurl.

    \subsection{Quantitative results}
    \label{sec:quantitative}

\paragraph{FAX improves faithfulness while preserving usability.}
    Table~\ref{tab:automatic_faithfulness} shows that FAX significantly outperforms all CRAFTER-XAI-Bench baselines in faithfulness.
    FAX achieves an average faithfulness score of 0.46 ($\pm$0.04 across runs), more than 2.3 times the 0.20 ($\pm$0.04) of the strongest baseline, a gap far larger than the run-to-run variation.
    At the same time, it maintains high Informativeness (0.90), Query Relevance (0.98), and Fluency (0.97).
    These usability scores are near the ceiling for all LLM-based methods: they separate natural-language responses from raw tool outputs, but fluent and relevant text does not by itself indicate faithfulness.

\paragraph{The verification stage prevents fluent but unsupported explanations.}
    The faithfulness of unstructured agentic XAI is slightly higher than that of Naive LLM, but the gap is not significant because the XAI methods themselves can be unfaithful.
    The low faithfulness of ExplainerDashboard is partly due to its low informativeness. Because our faithfulness metric is based on simulation, low informativeness makes simulation difficult.
    Structured Agentic XAI w/o Verification, an ablation of the verification stage, achieves the highest Informativeness (0.90), Query Relevance (0.99), and Fluency (0.99), but its faithfulness remains far lower than FAX. This result is central to our motivation: without verification, agentic systems are dangerously effective at producing articulate, relevant explanations that are fundamentally wrong.

\paragraph{The gain comes from claim verification, not from rewriting.}
    FAX adds both claim verification and a second synthesis pass, so its gain could come from rewriting alone.
    The Reflection-only variant in Table~\ref{tab:automatic_faithfulness} starts from the same plan, tool outputs, and draft as each FAX run, but replaces claim identification and testing with a generic reflection on which parts of the draft to keep, revise, or remove, followed by the same final synthesis (Appendix~\ref{appendix:reflection_only}).
    It reaches 0.18, close to Structured Agentic XAI w/o Verification (0.17) and far below FAX (0.46).
    Scoring all three variants together on identical drafts, FAX exceeds Reflection-only by 0.22 (95\% CI [0.16, 0.28]; higher in 34 of 41 scenarios), while Reflection-only and the unrevised draft differ by $-$0.002.
    The gain thus comes from testing specific claims against the target model, not from rewriting.

\paragraph{FAX transfers to tabular benchmarks.}
    Table~\ref{tab:tabular_results} reports results on the tabular datasets used in TalkToModel.
    FAX improves average faithfulness over TalkToModel on the original target models (0.70 vs. 0.65) and obtains substantially higher fluency (1.00 vs. 0.43), query relevance (0.86 vs. 0.34), and informativeness (0.65 vs. 0.26).
    These results indicate that explicit verification helps the agent translate noisy tabular XAI outputs into more complete and usable explanations.

\paragraph{Feature-ablated tabular results expose limitations in faithfulness assessment.}
    At the same time, the tabular setting clarifies an important evaluation limitation.
    Unlike CRAFTER-XAI-Bench, these tasks are binary classification problems over widely used datasets.
    The high faithfulness score of Naive LLM suggests that the backbone LLM can often predict model behavior from dataset-level knowledge and task priors, even without access to XAI tools.
    Therefore, a high simulation score in this setting does not necessarily indicate a faithful, model-specific explanation; it may instead reflect alignment between the LLM's prior and the learned classifier.
    The feature-ablated columns make this diagnostic explicit: we remove the most important feature of each original classifier (glucose, number of prior crimes, and loan amount), which domain knowledge also ties to the label, and retrain it, leaving the LLM's priors unchanged.
    Naive LLM drops from 0.74 to 0.68, driven by COMPAS (0.78 to 0.54), whereas FAX remains stable (0.70 to 0.71) and ties with Unstructured Agentic XAI for the highest faithfulness on the ablated models.
    This is consistent with prior sensitivity: methods that query the retrained model can follow its changed behavior, whereas Naive LLM relies on the same priors; as the effect varies across datasets and retraining also changes the model in other ways, we treat it as a diagnostic.
    The tabular benchmarks thus show that FAX transfers to the standard conversational XAI setting and why CRAFTER-XAI-Bench is needed as the primary benchmark for model-specific faithfulness.

\paragraph{Additional analyses.}
    Appendix~\ref{appendix:additional_quantitative_analyses} shows that a smaller Qwen3-8B evaluator gives lower absolute faithfulness but FAX remains the highest-scoring method, and that FAX's main cost is the additional token usage of verification.

    \subsection{Qualitative results}

    \paragraph{How each stage works.}
    Figure~\ref{fig:qualitative_verification_ablation} shows how the verification stage works.
    In the example, the response draft includes both claims inferred from SHAP explanations and additional claims based on the LLM’s domain knowledge.
    The agent tests these claims with state editing and rejects the unsupported ones in the final response.

    \paragraph{Model distinction emphasizes the necessity of CRAFTER-XAI-Bench.}
    In Figure~\ref{fig:analysis_model_distinction}, two target policies face the same state and choose the same action, but their learned decision rules differ.
    A plausible domain-level explanation is therefore insufficient; FAX produces explanations that separate the two policies, which is what CRAFTER-XAI-Bench tests: whether a response is grounded in the target model rather than in generic environment priors.
    
    \paragraph{Additional analysis on user specification. }
    CRAFTER-XAI-Bench also supports queries that specify the user's background or intended use (Appendix~\ref{appendix:user_specification_query}).

\section{Conclusions}

Faithful agentic XAI requires more than selecting and verbalizing XAI tool outputs. Because LLM agents can transform noisy evidence into fluent unsupported claims, explanation generation must include an explicit mechanism for checking which claims are actually grounded in the target model. FAX implements this idea by treating draft claims as hypotheses and testing them with model-executed interventions, and CRAFTER-XAI-Bench evaluates explanations in the same way, by checking the behavior they predict against target policies that cannot be explained by generic task knowledge. Our results show that this verification, rather than additional rewriting, substantially improves model-specific simulation faithfulness while preserving the usability benefits of natural-language interaction.

As the field of XAI continues to evolve and produce more diverse and sophisticated explanation methods, the importance of an agent that can critically evaluate, synthesize, and verify these outputs will only grow. FAX is a step toward agentic XAI systems that are not only accessible, but also faithful and trustworthy.

\section*{Acknowledgement}

This work was supported by the Institute of Information \& Communications Technology Planning \& Evaluation (IITP) grants funded by the Ministry of Science and ICT (MSIT) of the Republic of Korea: the Artificial Intelligence Graduate School Program (POSTECH) (RS-2019-II191906); the Information Technology Research Center (ITRC) Program (IITP-2026-RS-2024-00437866); the Global AI Frontier Lab International Collaborative Research Agreement (RS-2024-00469482 and RS-2024-00509258); the Development of Personalized Video Automatic Generation and Insertion Technology (RS-2026-25511821); and the AI Star Fellowship Support Program (IITP-2026-RS-2026-25616370). This work was also supported by a grant from KRAFTON AI. Jaechang acknowledges the POSTECH PIURI fellowship.

\section*{Limitations}

\paragraph{Computational overhead and latency}

A primary limitation of FAX is the increased computational cost and latency compared with standard non-agentic or unstructured agentic approaches.
Verification in FAX decomposes draft responses into claims, plans tests for those claims, executes additional faithful-tool calls, and synthesizes the resulting evidence.
As reported in Appendix~\ref{appendix:additional_quantitative_analyses}, FAX consumes about 38.7K tokens per question summed over all LLM calls, compared with 14.9K for Structured Agentic XAI w/o Verification and 6.0K for Unstructured Agentic XAI.
However, this cost is a direct trade-off for the central benefit of FAX: explanations become more faithful because unsupported claims are explicitly tested before the final response is generated.
For applications where faithful explanations matter, we view the additional computation as an acceptable cost for reducing the risk of fluent but unsupported explanations.

\paragraph{No human validation}
All metrics are computed automatically.
The faithfulness reference labels come from executing the target model, but the behavioral predictions are derived by an LLM evaluator, and we did not conduct a human simulation study in which participants predict model behavior from the explanations.
Such a study is needed to confirm that the measured gains translate into better human understanding.

\paragraph{Single backbone and evaluator family}
All agentic methods use the same backbone (Qwen3-32B on CRAFTER-XAI-Bench and Qwen3-8B on the tabular benchmarks) for a controlled comparison.
Because the quality of verification depends on the LLM's ability to design informative tests, the benefit of FAX may differ for other backbones, which we have not tested.
The cross-evaluator check in Appendix~\ref{appendix:additional_quantitative_analyses} also uses models from the same family.

\paragraph{Explanation-dependent tests}
The faithfulness score derives its tests from each explanation's claims, so different explanations are tested on different states.
An explanation that describes a few simple, directly tested behaviors may induce easier tests than one that describes a broader policy.
We report informativeness alongside faithfulness to expose this trade-off, but the metric does not fully control for test difficulty, and a benchmark with a fixed set of explanation-independent probe states would complement it.
Our ablation also does not include a control that gives a non-verifying agent the same number of additional model queries without targeting specific claims.

\paragraph{Scope and availability of faithful tools}
CRAFTER-XAI-Bench covers RL policies in a single environment, complemented by three tabular datasets.
FAX relies on tools whose outputs are tied to direct target-model execution, which requires query access to the model and an editable input representation.
State Editing and counterfactuals are examples of such tools for our settings; other domains need their own faithful tools, and where none are available, FAX can only mark claims as unverified rather than test them.

\section*{Ethical considerations}

FAX is intended to reduce a practical risk in agentic XAI: fluent natural-language explanations can make unreliable XAI outputs appear more trustworthy than they are.
By forcing the agent to verify claims against model-coupled evidence, FAX can help users identify unsupported explanations and can make XAI tools more accessible to non-expert users.
At the same time, the word ``verified'' should not be interpreted as a guarantee of global correctness.
FAX verifies local claims under available tools, states, and interventions; it can still miss relevant factors when faithful tools are unavailable, infeasible, or insufficiently informative.

This remaining uncertainty creates an over-trust risk.
In high-stakes domains such as health care, criminal justice, finance, and safety-critical control, users may reduce oversight if explanations are presented as authoritative.
Deployments in such settings require independent validation, human review, clear uncertainty communication, and domain-specific safety checks beyond the workflow evaluated in this paper.
There are also accessibility concerns: the additional token and tool-call cost may make verified agentic explanations harder to deploy for resource-constrained users, and adapting FAX to a new domain still requires expert effort to define valid faithful tools, state constraints, and evaluation scenarios.

\bibliography{custom}

\clearpage 
\appendix

\renewcommand{\thefigure}{A\arabic{figure}}
\setcounter{figure}{0}
\renewcommand{\thetable}{A\arabic{table}}
\setcounter{table}{0}

\section*{Appendix}
\label{sec:appendix}

\section{Extended Related Work}
\label{appendix:extended_related_work}

\subsection{Explainable AI}

\paragraph{Classical methods}
Post-hoc XAI methods include four broad families:
(i) \emph{feature attribution/saliency} that highlights input regions or features with high contribution~\citep{simonyan2014saliency};
(ii) \emph{surrogate models} that approximate a local/global decision rule~\citep{ribeiro2018anchors, ribeiro2016lime};
(iii) \emph{example-based explanations} such as prototypes and counterfactuals that reason via representative or minimally edited examples~\citep{chen2019protopnet, wachter2018counterfactual};
and (iv) \emph{concept-based explanations} that align internal representations with human-interpretable concepts~\citep{TCAV, yuksekgonul23posthoc}.
Each family exposes a different facet of model behavior,
and methods in the same family often produce different results~\cite{adebayo2018sanity}, which suggests potential unfaithfulness in explanations.
Consequently, a single method rarely satisfies diverse user intents.

\paragraph{Collection of explanations}
Since a single XAI method only reveals a limited aspect of a model's behavior, as illustrated in Figure~\ref{fig:xai_information_mapping}, frameworks like \citet{explainerdashboard, wenzhuo22omnixai, aix360} provide a collection of explanations in one place.
However, identifying which method best answers a user's question and how to interpret its output still requires nontrivial XAI/ML expertise.
In practice, users face a selection and interpretation burden: they must map their intent to a suitable method and often combine multiple views.

\paragraph{Interactive XAI}
To lower the barrier for non-experts, recent works have focused on generating natural language explanations that verbalize XAI outputs~\citep{Zytek24explingo, castelnovo2024augmenting}.
Conversational assistants have been proposed to explain a model's reasoning to users~\citep{Zhang25followup}, and the benefits of text-based explanations over classical methods have been confirmed in human studies~\citep{lakkaraju2022rethinking, mindlin2024measuring}.
Building on this, \emph{Agentic XAI} systems have emerged, which use LLMs to select appropriate XAI tools based on a user's query~\citep{slack23talktomodel, he2025conversational}.
In a broader vision, \citet{kim2025agentic} advocate \emph{agentic interpretability}, a multi-turn conversation in which an LLM proactively helps users understand it by building and using a mental model of the user.
Whereas agentic interpretability concerns an LLM explaining itself, Agentic XAI uses an LLM to explain a separate target model through XAI tools.

However, these pioneering agentic systems have two critical limitations. First, they have primarily been tested on simpler models in static, tabular data settings. Second, and more crucially, they implicitly assume the underlying XAI tools are consistently faithful. This assumption often breaks down in complex and dynamic environments, where the unfaithfulness of XAI methods is a known and severe issue~\citep{adebayo2018sanity}. An agent that naively trusts and translates unreliable tool outputs can produce fluent, plausible, yet fundamentally incorrect explanations. \citet{he2025conversational} have also warned that LLMs may amplify users' misunderstandings, and \citet{kim2025agentic} note that agentic interpretability trades off completeness for interactivity, making it less suited to high-stakes settings. We address this critical gap by focusing on enhancing explanation faithfulness within a challenging, dynamic environment.

\subsection{LLM agent and agentic workflow}
Recent work frames LLMs as \emph{agents} that plan, act, and reflect, including through the use of external tools.
ReAct interleaves reasoning traces with environment-facing actions to update plans and handle exceptions~\citep{yao2022react}, while Toolformer demonstrates that LMs can \emph{self-learn} when and how to call APIs and integrate their outputs~\citep{schick2023toolformer}.
Building on these foundations, agentic extensions of LLMs now emphasize workflows that support multi-step reasoning, memory, and adaptive decision-making.
For instance, the Model Context Protocol (MCP) provides a standardized interface for connecting LLMs with external services and tools, enabling modular extensibility.
In contrast to unstructured workflows, which allow the LLM to decide plans and actions dynamically,
recent work emphasizes that structured workflows are essential for reliable and stable orchestration of agent behaviors on specific tasks~\citep{zhang2025aflow}.
These developments underscore that the design of robust agentic workflows is central to realizing LLMs as proactive agents capable of simulation, decision-making, and long-horizon interaction.

\subsection{Scalable evaluation of generated texts and explanations}

LLM judges have emerged as a practical, scalable proxy for costly human studies, especially for evaluating the quality of generated text.
\citet{zheng2023judging} demonstrated that strong LLM judges can achieve high agreement with human preferences.
Rubric-driven evaluators like G-Eval further improve human alignment by leveraging chain-of-thought and structured outputs~\citep{geval}.

For evaluating explanations,
faithfulness has been evaluated through \emph{simulatability}: the degree to which an explanation helps an observer predict the model's behavior on unseen inputs~\citep{lyu2024towards}.
The underlying assumption is that a faithful explanation should allow one to reproduce the model's decision-making process~\citep{jacovi2020towards}.
Prior work has implemented this idea by training student models~\citep{li2020evaluating} or by asking humans to act as simulators~\citep{chen2018learning, nguyen2018comparing, hase2020evaluating}.
In contrast, we employ an LLM as a simulator. After observing an input, the model's output, and the corresponding explanation, the LLM is tasked with predicting the model's behavior in new, unseen situations.
By comparing the LLM's simulated predictions with the model's actual outputs, we compute a simulation accuracy score, which serves as our quantitative measure of faithfulness.

\section{Detailed Definition of Inherently Faithful Tools}
\label{appendix:inherently_faithful_tools}

This section provides a detailed definition of \emph{inherently faithful tools}.
The definition is centered on the explanation claim produced by the tool, not on whether the tool produces a complete account of the model's internal mechanism.
An inherently faithful tool produces evidence for which the central explanatory statement can be judged true or false by directly executing the target model, and an explanation generated from that tool uses this model-executed result as the explanation evidence.

\paragraph{Operational definition}
Let $\pi$ denote the target model and $s$ denote a concrete input state.
An explanation claim $c$ is directly model-verifiable when there exists a concrete model query whose execution determines whether $c$ is true.
For example, the claim ``under edited state $s'$, the target model chooses action $a$'' is true exactly when $a=\pi(s')$.
A tool is inherently faithful when its explanation evidence has this form and the tool reports the result of the target-model execution itself, such as $\pi(s)$, $\pi(s')$, or whether $\pi(s')$ satisfies a requested decision condition.
In FAX, a final explanation is faithful with respect to such a tool when it directly preserves the verified result: for instance, it may state that the target model chooses $a$ after a specified edit, or that a proposed counterfactual is valid because executing the target model on the counterfactual state changes the decision as required.

This definition differs from explanations that are inferred from intermediate quantities.
Sanity-check work shows that some explanation methods can produce similar outputs even when model parameters or labels are changed~\citep{adebayo2018sanity}.
Such issues can arise when explanation claims are based on gradients, saliency maps, perturbation scores, surrogate models, background distributions, or other approximations rather than on the model decision that the explanation claims to describe.
These methods can still be useful and informative, but their outputs are potentially unfaithful evidence under our definition because the truth of the explanation claim is not directly determined by executing the target model.

\paragraph{State Editing}
State Editing is an intervention tool: given an observed state $s_t$ and an edit instruction $\delta$, it constructs an edited state $s'_t=\mathrm{edit}(s_t,\delta)$ and directly executes the target policy to obtain $\pi(s'_t)$.
Its explanation evidence is therefore aligned with the model decision by construction: it does not estimate what the model would do after the edit, but reports the model's actual decision under the edited state.
FAX uses this evidence to test claims such as whether changing a resource, object, or event changes the action in the expected direction.
However, State Editing remains local and intervention-specific; invalid edits, impossible states, or edits that change multiple semantic factors should be marked inconclusive, and even valid results may be less informative than broader post-hoc explanations.

\paragraph{Counterfactuals}
Counterfactual explanation methods ask for an alternative input or state $s_{\mathrm{cf}}$ that is close to the original state $s_t$ but changes the target model decision, e.g., $\pi(s_{\mathrm{cf}}) \neq \pi(s_t)$ or $\pi(s_{\mathrm{cf}})=a^*$ for a requested target action $a^*$~\citep{wachter2018counterfactual, karimi2020mace}.
In our setting, a counterfactual tool is inherently faithful when candidate counterfactuals are accepted only after directly executing the target model on the candidate state and confirming that the required decision change occurs.
The search procedure may use heuristics or optimization, but the explanation claim ``this edit is a valid counterfactual'' is grounded in the target model's actual output, not in an approximation of what might change the model.
At the same time, an accepted counterfactual shows only the verified decision-changing relation under the chosen search space, distance metric, feasibility constraints, and state representation; it does not by itself prove a unique cause or a globally stable rule.

\paragraph{Trade-off with richer post-hoc tools}
The phrase ``inherently faithful'' therefore marks a trade-off, not an absolute superiority claim.
Tools such as SHAP and HIGHLIGHTS can summarize broad patterns, highlight salient features or time steps, and provide information that direct interventions may not easily expose.
However, because they rely on attribution, saliency, sampling, or other intermediate calculations, their outputs can be potentially unfaithful to the target model decision and should not be treated as decisive evidence without verification.
Inherently faithful tools avoid this particular source of unfaithfulness by tying validity to model execution, but they often communicate less information: they usually answer local questions such as ``what does the model do under this edited state?'' or ``does this candidate edit change the decision?'' rather than producing a full explanatory narrative.
FAX uses this complementarity deliberately: richer post-hoc tools help generate candidate claims, and inherently faithful tools test whether those claims survive direct alignment with the target model's decisions.

\section{Detailed Baseline Descriptions}
\label{appendix:baseline_descriptions}

We compare FAX against the following baselines.
\begin{itemize}
    \item ExplainerDashboard~\citep{explainerdashboard}: A non-agentic approach where results from multiple XAI tools are collected and presented directly. In CRAFTER-XAI-Bench, it uses the same XAI tools except State Editing, which requires a concrete edit instruction and is therefore unavailable to a non-interactive baseline.
    \item TalkToModel~\citep{slack23talktomodel}: A conversational XAI baseline that maps user queries to tool calls through a parser and returns explanations from the selected tool outputs. Because its operation grammar is defined for tabular models, we use TalkToModel as the reference baseline in the tabular setting.
    \item Naive LLM: A baseline that generates explanations without access to XAI tools, relying on the backbone LLM and the domain knowledge provided in the system prompt. This tests whether apparent faithfulness can come from generic task knowledge rather than model-grounded evidence.
    \item Unstructured Agentic XAI: An agent that can use XAI tools freely without a predefined workflow. It can issue several tool calls before answering, but it is not explicitly required to verify claims. This baseline, inspired by~\citep{he2025conversational}, tests the value of a structured workflow.
    \item Structured Agentic XAI w/o Verification: A direct ablation of FAX that follows the same planning, tool execution, and drafting workflow but omits the verification and synthesis stage. This isolates the contribution of explicit claim verification.
\end{itemize}

\section{Implementation Details}
\label{appendix:implementation_details}

\paragraph{Target policies}
The three CRAFTER-XAI-Bench policies are DQN agents with a residual MLP Q-network (four layers of 512 units).
They act on a symbolic Crafter observation of the $9\times7$ local map, inventory, vital statistics, facing direction, light level, sleeping flag, and achievements; State Editing exposes these as 104 named features.
Each policy is explained on states sampled from three held-out test episodes, and the inputs given to every method are a textual description of the state, the chosen action, and the relative Q-values of the top actions.
We release the trained checkpoints but not the training code.

\paragraph{XAI tools}
SHAP and MACE are computed with OmniXAI~\citep{wenzhuo22omnixai} over the symbolic observation, using episodes collected from the policies as background data; MACE counterfactuals for the benchmark states are precomputed and cached.
The episode summary follows HIGHLIGHTS~\citep{amir2018highlights}: it reports the five most important time steps of the current episode by Q-value importance, with temporal non-maximum suppression and the status and inventory changes around each step.
State Editing applies the requested feature edits to the symbolic observation and returns the target policy's action and Q-values on the edited state.
All tools run in a separate tool server queried over HTTP.

\paragraph{Workflows}
FAX is a LangGraph graph with four LLM stages.
In the planning stage, the LLM may call any of the four tools (with parallel tool calls); the drafting stage writes $E_{draft}$ without tools; in the verification stage, the LLM lists the draft's claims and may call State Editing and the counterfactual tool, with up to three calls per claim; the final stage writes $E_{final}$ from the full message history.
Structured Agentic XAI w/o Verification runs the same graph up to the draft.
Unstructured Agentic XAI issues one LLM turn with access to all tools, executes the requested calls, and answers in a second turn.
Naive LLM answers in a single call without tools.
For FAX and the baselines, sampling parameters are left at the serving defaults with Qwen3's thinking mode enabled, and reasoning traces are removed before evaluation.

\paragraph{Evaluation}
For faithfulness, the evaluator LLM generates five hypotheses per explanation in JSON (Figure~\ref{fig:appendix_system_prompt_faith}), each state edit is executed with State Editing on the scenario state, and a hypothesis counts as correct if the predicted action equals the policy's action.
For informativeness, query relevance, and fluency, we compute the probability-weighted average of the evaluator's scores from 1 to 5 using token log-probabilities, with thinking disabled, and rescale it to $[0,1]$ as $(x-1)/4$.

\section{Tabular Benchmark Setup and Full Results}
\label{appendix:tabular_full}

\paragraph{Datasets and models}
We use the Diabetes, COMPAS, and German Credit datasets and the gradient-boosted tree classifiers released with TalkToModel~\citep{slack23talktomodel}, and explain each classifier on instances from its test split.
All agentic methods access the model through TalkToModel's XAI operations (e.g., feature importance, what-if edits, counterfactuals, prediction likelihood, and data and model statistics); the editing and counterfactual operations play the role of the inherently faithful tools.

\paragraph{Feature-ablated models}
For each dataset, we remove the feature with the highest importance in the original classifier (Diabetes: glucose; COMPAS: number of prior crimes; German Credit: loan amount) and retrain a classifier of the same type with the same training recipe and random seed on the remaining features.
These features are also the ones that domain knowledge most directly associates with the label, so the LLM's priors still point to a feature that the ablated model can no longer use.

\paragraph{Queries and runs}
Each dataset has 12 queries, three for each of four types: feature importance, counterfactual, what-if, and model understanding.
All LLM-based methods and the evaluator use Qwen3-8B.
Each cell is the mean of two replications, each with two or three generations per query.
In these runs, the Naive LLM and Unstructured Agentic XAI prompts did not include the dataset description, which the other methods received.

\paragraph{Faithfulness}
The evaluator derives five hypotheses from an explanation, each an edit of the instance together with the class the explanation implies.
Because the target is binary, a hypothesis is counted as correct if the edit moves the model's predicted probability of the positive class in the implied direction; edits that change the probability by less than 0.01 receive half credit, and hypotheses that copy an edit stated verbatim in the explanation are down-weighted to 0.3 because they do not test the evaluator's reading of the explanation's claims.
A random explanation is therefore expected to score about 0.5.

Table~\ref{tab:tabular_results_full} provides the full results on tabular data.

\begin{table*}[t]
\centering
\caption{
    Evaluation on the tabular benchmarks used by TalkToModel, under the original model and a feature-ablated model variant in which the top prior-aligned feature is removed before retraining.
    This setting tests FAX in the benchmark family used by prior conversational XAI systems, while the ablated-model columns diagnose prior sensitivity in tabular faithfulness evaluation.
    All metrics are rescaled to 0-1.
    Note that the expected faithfulness of a random explanation is 0.5 because it is binary classification.
}
\label{tab:tabular_results_full}
\resizebox{0.9\textwidth}{!}{%
\begin{tabular}{clcccccccc}
\toprule
 & & \multicolumn{4}{c}{Original Model} & \multicolumn{4}{c}{Feature-Ablated Model} \\
\cmidrule(lr){3-6} \cmidrule(lr){7-10}
Method & Dataset & Faithful. & Informat. & Query Rel. &  Fluency  & Faithful.& Informat. & Query Rel. &  Fluency \\
\midrule

\multirow{4}{*}{\makecell{Explainer \\Dashboard}}
& diabetes & 0.55 &   0.51 & 0.51 & 0.42 & 0.58 &  0.44 & 0.50 & 0.58  \\
& compas   & 0.50 &  0.49 & 0.51 &  0.42 & 0.64 &  0.47 & 0.52 & 0.43  \\
& german   & 0.66 &   0.38 & 0.47 & 0.52 & 0.80 &  0.39 & 0.42 & 0.60  \\
\cmidrule{2-10}
& average  & 0.57 &   0.46 & 0.49 & 0.45
           & 0.67 &  0.43 & 0.48 & 0.54  \\
\midrule

\multirow{4}{*}{TalkToModel}
& diabetes & 0.66 & 0.26   & 0.41 &0.45 & 0.67 &  0.22 & 0.30 & 0.47 \\
& compas   & 0.67 & 0.25   & 0.34 &0.45 & 0.68 &  0.24 & 0.32 & 0.44 \\
& german   & 0.62 & 0.26   & 0.28 &0.38 & 0.60 &  0.19 & 0.21 & 0.43 \\
\cmidrule{2-10}
& average  & 0.65 & 0.26 & 0.34 & 0.43
           & 0.65 & 0.21 & 0.28 & 0.45 \\
\midrule

\multirow{4}{*}{Naive LLM}
& diabetes & 0.71 & 0.78   & 0.87 &1.00 & 0.79 &  0.79 & 0.80 & 1.00 \\
& compas   & 0.78 & 0.77   & 0.91 &0.99 & 0.44 &  0.70 & 0.88 & 0.98 \\
& german   & 0.74 & 0.87   & 0.89 &0.99 & 0.66 &  0.91 & 0.87 & 0.99 \\
\cmidrule{2-10}
& average  & \textbf{0.74} &\textbf{0.81}  & \textbf{0.89} &0.99 
           & 0.60 &\textbf{0.80} & 0.85 &0.99   \\
\midrule

\multirow{4}{*}{\makecell[c]{ Unstructured \\ Agentic XAI }}
& diabetes & 0.75 &  0.62 & 0.70 & 0.88 & 0.80 & 0.58 & 0.69 & 0.96 \\
& compas   & 0.73 &  0.49 & 0.62 & 0.81 & 0.55 & 0.65 & 0.77 & 0.94 \\
& german   & 0.56 &  0.62 & 0.73 & 0.90 & 0.65 & 0.46 & 0.62 & 0.78 \\
\cmidrule{2-10}
& average  & 0.68 &  0.58 & 0.68 & 0.86 & 0.67 & 0.56 & 0.69 & 0.89 \\
\midrule

\multirow{4}{*}{\makecell[c]{Structured \\ Agentic XAI \\ w/o verification}}
& diabetes & 0.81 &  0.71 & 0.78 & 1.00 & 0.77 & 0.70 & 0.83 & 1.00 \\
& compas   & 0.71 &  0.71 & 0.82 & 0.99 & 0.64 & 0.66 & 0.87 & 0.98 \\
& german   & 0.61 &  0.72 & 0.87 & 0.99 & 0.60 & 0.75 & 0.85 & 0.98 \\
\cmidrule{2-10}
& average  & 0.71 &  0.71 & 0.82 & 0.99 & 0.67 & 0.71 & 0.85 & 0.99 \\
\midrule

\multirow{4}{*}{FAX (proposed)}
& diabetes & 0.79 & 0.68   & 0.86 &1.00 & 0.79 &  0.70 & 0.90 & 1.00 \\
& compas   & 0.69 & 0.62   & 0.84 &1.00 & 0.67 &  0.68 & 0.96 & 0.99 \\
& german   & 0.63 & 0.65   & 0.87 &1.00 & 0.66 &  0.77 & 0.86 & 1.00 \\
\cmidrule{2-10}
& average  & 0.70 &0.65 & 0.86 & \textbf{1.00}
           & \textbf{0.71} & 0.72 & \textbf{0.91} & \textbf{1.00} \\

\bottomrule
\end{tabular}
}
\end{table*}

\section{System prompts for Agentic XAI methods}
\label{appendix:prompt_xai}
Figures~\ref{fig:appendix_system_prompt_plan}, \ref{fig:appendix_system_prompt_draft}, \ref{fig:appendix_system_prompt_verification}, and \ref{fig:appendix_system_prompt_final} illustrate the full system prompts employed in FAX.

\begin{figure*}[h]
    \centering
    \begin{tcolorbox}[title={\textbf{\small Implementation prompt}}, boxrule=2pt, arc=0mm]\begin{minted}[fontsize=\scriptsize, breaklines, breakanywhere, frame=lines, framesep=2mm, tabsize=4, style=vs, autogobble]{text}
You are a helpful explanation curator for a model in a 2d Minecraft-like game called 'crafter'.
Note that the model have its own (unknown) goals, so do not regard it based on a stereotype of typical behavior.
You have access to tools to get XAI explanations or predictions.

Your task is to answer the user's question by following a strict workflow.
This is the FIRST step: PLAN.

**Environment description:** {ENVIRONMENT_DESCRIPTION}
**User's Question:** {USER_QUESTION}
**Initial State & Model Decision:**
{STATE_DESCRIPTION_AND_MODEL_DECISION}

Based on the user's question and the initial state, create a plan.
Decide which tools you need to call to gather the necessary information.
Then, call those tools.

\end{minted}
\end{tcolorbox}
        \caption{
            System prompt for the planning stage in FAX on CRAFTER-XAI-Bench (verbatim; placeholders in braces).
        }
        \label{fig:appendix_system_prompt_plan}
\end{figure*}

\begin{figure*}[h]
    \centering
    \begin{tcolorbox}[title={\textbf{\small Implementation prompt}}, boxrule=2pt, arc=0mm]\begin{minted}[fontsize=\scriptsize, breaklines, breakanywhere, frame=lines, framesep=2mm, tabsize=4, style=vs, autogobble]{text}
This is RESPONSE GENERATION step.
You have completed all information gathering.
Using all the information from the previous steps, write a comprehensive final response to the user's original question.

**User's Original Question:** {USER_QUESTION}
**Tool Results:**
{TOOL_RESULTS}

Structure your answer clearly, using the explanations as supporting materials.

\end{minted}
\end{tcolorbox}
        \caption{
            System prompt for the draft generation stage in FAX (verbatim; placeholders in braces).
        }
        \label{fig:appendix_system_prompt_draft}
\end{figure*}

\begin{figure*}[h]
    \centering
    \begin{tcolorbox}[title={\textbf{\small Implementation prompt}}, boxrule=2pt, arc=0mm]\begin{minted}[fontsize=\scriptsize, breaklines, breakanywhere, frame=lines, framesep=2mm, tabsize=4, style=vs, autogobble]{text}
This is the intermediate step: VERIFICATION.
You have executed your initial plan, received tool results, and generated a response draft. Your objective is to verify whether the claims in the response draft faithfully reflect the model's actual behavior.
Perform the following steps:
1. Claim Identification
- Identify all substantive claims in the response draft that are relevant to understanding the model and answering the user's question.
- Decompose complex claims into atomic, independently verifiable claims.
- If there are no claims to verify, state "Verification is not needed." and do not call any tools.
2. Evidence Analysis
- For each identified claim, examine the existing tool results and identify supporting, contradictory, or missing evidence.
- Assess whether each claim is fully supported, partially supported, unsupported, or contradicted by the available evidence.
- Recall that SHAP and Episode Summary can be noisy, whereas state editing and counterfactual analysis provide faithful evidence of model behavior.
- Do not treat feature attribution or observational patterns alone as sufficient evidence for causal claims.
3. Falsification Planning
- For each claim, formulate a verification plan using the edit_state tool.
- Design interventions that can both support and potentially falsify the claim, rather than merely confirm it.
- For each planned tool call, specify:
-- The state variables to modify and the intervention to perform.
-- The expected model behavior if the claim is true.
-- The observable outcome that would contradict or weaken the claim.
- Where possible, isolate the relevant variables and control for alternative explanations.
- Prioritize informative interventions that distinguish the claim from competing hypotheses.
- You may plan up to three tool calls per claim.
4. Verification Execution
- Execute the planned tool calls, using no more than three calls per claim.
- Compare the observed model behavior with the expected supporting and falsifying outcomes specified in the verification plan.
- Evaluate both confirming and disconfirming evidence without confirmation bias.
- Update the verification status of each claim as Supported, Partially Supported, Falsified, or Inconclusive.
- Identify claims that need to be retained, revised, qualified, or removed from the response draft.
- Do not consider a claim conclusively verified merely because a limited set of interventions failed to falsify it.
- Base your verification conclusions on faithful tool results rather than assumptions or the original response draft.

\end{minted}
\end{tcolorbox}
        \caption{
            System prompt for the verification stage in FAX on CRAFTER-XAI-Bench (verbatim). In this stage, only the inherently faithful tools are available to the agent.
        }
        \label{fig:appendix_system_prompt_verification}
\end{figure*}

\begin{figure*}[h]
    \centering
    \begin{tcolorbox}[title={\textbf{\small Implementation prompt}}, boxrule=2pt, arc=0mm]\begin{minted}[fontsize=\scriptsize, breaklines, breakanywhere, frame=lines, framesep=2mm, tabsize=4, style=vs, autogobble]{text}
This is the FINAL step: FINAL RESPONSE.
You have completed all information gathering and verification.
Using all the information from the previous steps, write a comprehensive final response to the user's original question.

**User's Original Question:** {USER_QUESTION}
**Initial Plan & Tool Execution Results:**
(Contained in the message history)
{VERIFICATION_TOOL_RESULTS}

Structure your final answer clearly, using the explanations as supporting materials. Be conservative with any conjectures.

\end{minted}
\end{tcolorbox}
        \caption{
            System prompt for the final response generation stage in FAX (verbatim; placeholders in braces).
        }
        \label{fig:appendix_system_prompt_final}
\end{figure*}

\section{Reflection-only Ablation}
\label{appendix:reflection_only}

\paragraph{Protocol}
The ablation isolates claim verification from rewriting by holding everything before the verification stage fixed.
For each of the 123 FAX runs (41 scenarios $\times$ three runs), we take the logged plan, tool outputs, and draft, and compare three continuations:
FAX, the logged verification and final response;
Reflection-only, a generic reflection turn (Figure~\ref{fig:appendix_system_prompt_reflection_only}) with no claim identification and no tool access, followed by the FAX final-response prompt;
and the unrevised draft, which is what Structured Agentic XAI w/o Verification returns for the same head.
In all 123 runs, the reflection turn made no tool calls.
The three variants were scored together in a separate evaluation run with the same evaluator model and faithfulness protocol as Table~\ref{tab:automatic_faithfulness}; in this run, usability scores are the evaluator's integer scores rather than probability-weighted averages.
Because the evaluator samples its hypotheses, absolute scores differ between evaluation runs (e.g., FAX scores 0.40 here and 0.46 in Table~\ref{tab:automatic_faithfulness}), so we compare the variants within this run.
The Reflection-only rows of Table~\ref{tab:automatic_faithfulness} are taken from this run.

\paragraph{Results}
Table~\ref{tab:reflection_only} reports faithfulness and paired differences over scenarios, with 95\% bootstrap confidence intervals (10{,}000 resamples) and two-sided Wilcoxon signed-rank tests.

\begin{table}[h]
\centering
\small
\caption{Reflection-only ablation: faithfulness of the three continuations of the same FAX runs, scored in one evaluation run (mean over 41 scenarios; $\pm$: standard deviation across the three runs), and paired differences over scenarios.}
\label{tab:reflection_only}
\resizebox{\linewidth}{!}{%
\begin{tabular}{lc}
\toprule
\textbf{Variant} & \textbf{Faithfulness} \\
\midrule
FAX & 0.40{\scriptsize$\pm$0.02} \\
Reflection-only & 0.18{\scriptsize$\pm$0.03} \\
Unrevised draft (w/o verification) & 0.18{\scriptsize$\pm$0.01} \\
\midrule
\textbf{Paired difference} & \textbf{$\Delta$ [95\% CI], $p$, win/loss/tie} \\
\midrule
FAX $-$ Reflection-only & $+$0.22 [0.16, 0.28], $p<10^{-4}$, 34/4/3 \\
Reflection-only $-$ draft & $-$0.00 [$-$0.05, 0.05], $p=0.77$, 15/16/10 \\
FAX $-$ draft & $+$0.22 [0.15, 0.28], $p<10^{-4}$, 29/5/7 \\
\bottomrule
\end{tabular}
}
\end{table}

\begin{figure*}[h]
    \centering
    \begin{tcolorbox}[title={\textbf{\small Implementation prompt}}, boxrule=2pt, arc=0mm]\begin{minted}[fontsize=\scriptsize, breaklines, breakanywhere, frame=lines, framesep=2mm, tabsize=4, style=vs, autogobble]{text}
This is the intermediate step: REFLECTION.
You have executed your initial plan and received the following tool results, and generated response draft.

Now, reflect on the response draft.
- Reflect on the response draft for understanding the model and answering the user's question.
- Consider which parts of the response should be kept, revised, or removed.
- If there is nothing to reflect on, state 'Reflection is not needed.'

\end{minted}
\end{tcolorbox}
        \caption{
            Reflection prompt of the Reflection-only ablation (verbatim). It is derived from the FAX verification prompt (Figure~\ref{fig:appendix_system_prompt_verification}) by removing claim identification, evidence checking, and tool planning. The final-response prompt is the FAX prompt (Figure~\ref{fig:appendix_system_prompt_final}) with ``verification'' replaced by ``reflection''.
        }
        \label{fig:appendix_system_prompt_reflection_only}
\end{figure*}

\section{Full user query list}
\label{appendix:query_list}
Table~\ref{tab:scenarios} provides the complete list of user queries used for evaluation.

\begin{table*}[h]
    \centering
    \caption{The 41 evaluation scenarios in CRAFTER-XAI-Bench (10 plan, 10 why, 11 what-if, and 10 counterfactual).}
    \label{tab:scenarios}
    
        \resizebox{\textwidth}{!}{%
        \begin{tabular}{l l l c}
        \toprule
        \textbf{Category} & \textbf{Query} & \textbf{Model} & \textbf{State ID} \\
        \midrule
        \multirow{10}{*}{Plan}
        & \multirow{6}{*}{What is the model's immediate plan?} & diamond & diamond\_60 \\
        & & diamond & diamond\_67 \\
        & & diamond & diamond\_330 \\
        & & hoarder & hoarder\_160 \\
        & & hoarder & hoarder\_302 \\
        & & pacifist & pacifist\_110 \\
        \cmidrule(lr){2-4}
        & \multirow{4}{*}{What is the model's future plan?} & diamond & diamond\_101 \\
        & & hoarder & hoarder\_302 \\
        & & pacifist & pacifist\_50 \\
        & & pacifist & pacifist\_741 \\
        \midrule
        \multirow{10}{*}{Why}
        & \multirow{3}{*}{Why does the model collect wood?} & diamond & diamond\_60 \\
        & & hoarder & hoarder\_161 \\
        & & pacifist & pacifist\_50 \\
        \cmidrule(lr){2-4}
        & \multirow{3}{*}{Why does the model craft a pickaxe instead of a sword?} & diamond & diamond\_67 \\
        & & hoarder & hoarder\_10 \\
        & & pacifist & pacifist\_741 \\
        \cmidrule(lr){2-4}
        & \multirow{4}{*}{Why does the model not run away from monsters?} & diamond & diamond\_101 \\
        & & hoarder & hoarder\_120 \\
        & & pacifist & pacifist\_50 \\
        & & pacifist & pacifist\_680 \\
        \midrule
        \multirow{11}{*}{What if}
        & \multirow{4}{*}{Does the model change its action if its inventory is empty?} & diamond & diamond\_60 \\
        & & diamond & diamond\_330 \\
        & & hoarder & hoarder\_302 \\
        & & pacifist & pacifist\_110 \\
        \cmidrule(lr){2-4}
        & \multirow{3}{*}{Would the model change its plan if the model knew where a diamond is?} & diamond & diamond\_60 \\
        & & hoarder & hoarder\_302 \\
        & & pacifist & pacifist\_110 \\
        \cmidrule(lr){2-4}
        & \multirow{4}{*}{If a wood pickaxe disappears from inventory, will the model craft it again?} & diamond & diamond\_101 \\
        & & hoarder & hoarder\_302 \\
        & & pacifist & pacifist\_442 \\
        & & pacifist & pacifist\_741 \\
        \midrule
        \multirow{10}{*}{Counterfactual}
        & \multirow{6}{*}{When does the model attack a monster?} & diamond & diamond\_101 \\
        & & hoarder & hoarder\_120 \\
        & & hoarder & hoarder\_302 \\
        & & pacifist & pacifist\_442 \\
        & & pacifist & pacifist\_680 \\
        & & pacifist & pacifist\_741 \\
        \cmidrule(lr){2-4}
        & \multirow{4}{*}{When will the model sleep?} & diamond & diamond\_60 \\
        & & diamond & diamond\_101 \\
        & & diamond & diamond\_330 \\
        & & hoarder & hoarder\_160 \\
        \bottomrule
        \end{tabular}%
        }
    
\end{table*}

\clearpage

\section{Additional Quantitative Analyses}
\label{appendix:additional_quantitative_analyses}

\paragraph{LLM Judge robustness.}
Because the faithfulness metric uses an LLM evaluator, we check whether the relative result is sensitive to the judge model.
Table~\ref{tab:judge_robustness} compares the primary Qwen3-32B evaluator with Qwen3-8B.
The smaller judge produces lower absolute scores, which is expected because simulation requires non-trivial reasoning over generated states.
However, FAX remains the highest-scoring method under both evaluators, and the cross-judge Intraclass Correlation Coefficient is 0.46, indicating moderate agreement.

\begin{table}[t]
\centering
\small
\caption{
    Cross-model robustness of the faithfulness evaluator.
    FAX remains the highest-scoring method under both the primary Qwen3-32B judge and a smaller Qwen3-8B judge.
}
\label{tab:judge_robustness}
\resizebox{\linewidth}{!}{%
\begin{tabular}{lcc}
\toprule
\textbf{Method} & \textbf{Qwen3-32B} & \textbf{Qwen3-8B} \\
\midrule
Explainer Dashboard & 0.20 & 0.17 \\
Naive LLM & 0.14 & 0.06 \\
Unstructured Agentic XAI & 0.18 & 0.09 \\
Structured Agentic XAI w/o Verification & 0.17 & 0.16 \\
FAX (ours) & \textbf{0.46} & \textbf{0.28} \\
\bottomrule
\end{tabular}
}
\end{table}

\paragraph{Token usage.}
Table~\ref{tab:token_usage} reports the average token usage per question with two statistics: the largest single LLM call and the sum over all calls.
FAX requires more tokens than the baselines because it explicitly decomposes draft responses into claims, plans interventions, verifies those claims, and synthesizes the resulting evidence.
Summed over all calls, FAX uses about 2.6 times the tokens of Structured Agentic XAI w/o Verification and 6.5 times those of Unstructured Agentic XAI, because every additional stage re-sends the growing conversation history.
This overhead is the main computational cost of verification in FAX.
We report token usage as a model-independent cost indicator; wall-clock latency and monetary cost can vary substantially with serving hardware and deployment settings.

\begin{table}[t]
\centering
\small
\caption{
    Token usage per question on CRAFTER-XAI-Bench (mean over 41 scenarios and three runs).
    \textit{Largest call}: prompt and completion tokens of the largest single LLM call, which approximates the final context length.
    \textit{All calls}: tokens summed over every LLM call, i.e., total consumption; agentic methods re-send the conversation history at each call.
    $^{*}$Naive LLM makes a single call whose usage was not logged; we count its prompt tokens with the Qwen3 tokenizer.
}
\label{tab:token_usage}
\resizebox{\linewidth}{!}{%
\begin{tabular}{lcc}
\toprule
\textbf{Method} & \textbf{Largest call} & \textbf{All calls} \\
\midrule
Naive LLM & 506$^{*}$ & 506$^{*}$ \\
Unstructured Agentic XAI & 2{,}979 & 5{,}958 \\
Structured Agentic XAI w/o Verification & 3{,}848 & 14{,}855 \\
FAX (ours) & 6{,}117 & 38{,}663 \\
\bottomrule
\end{tabular}
}
\end{table}

\section{Additional Agentic XAI Scenarios in Crafter}
\label{appendix:additional_crafter_scenarios}

\subsection{User specification in query}
\label{appendix:user_specification_query}

\begin{figure*}[t]
    \centering
    \includegraphics[width=0.9\linewidth]{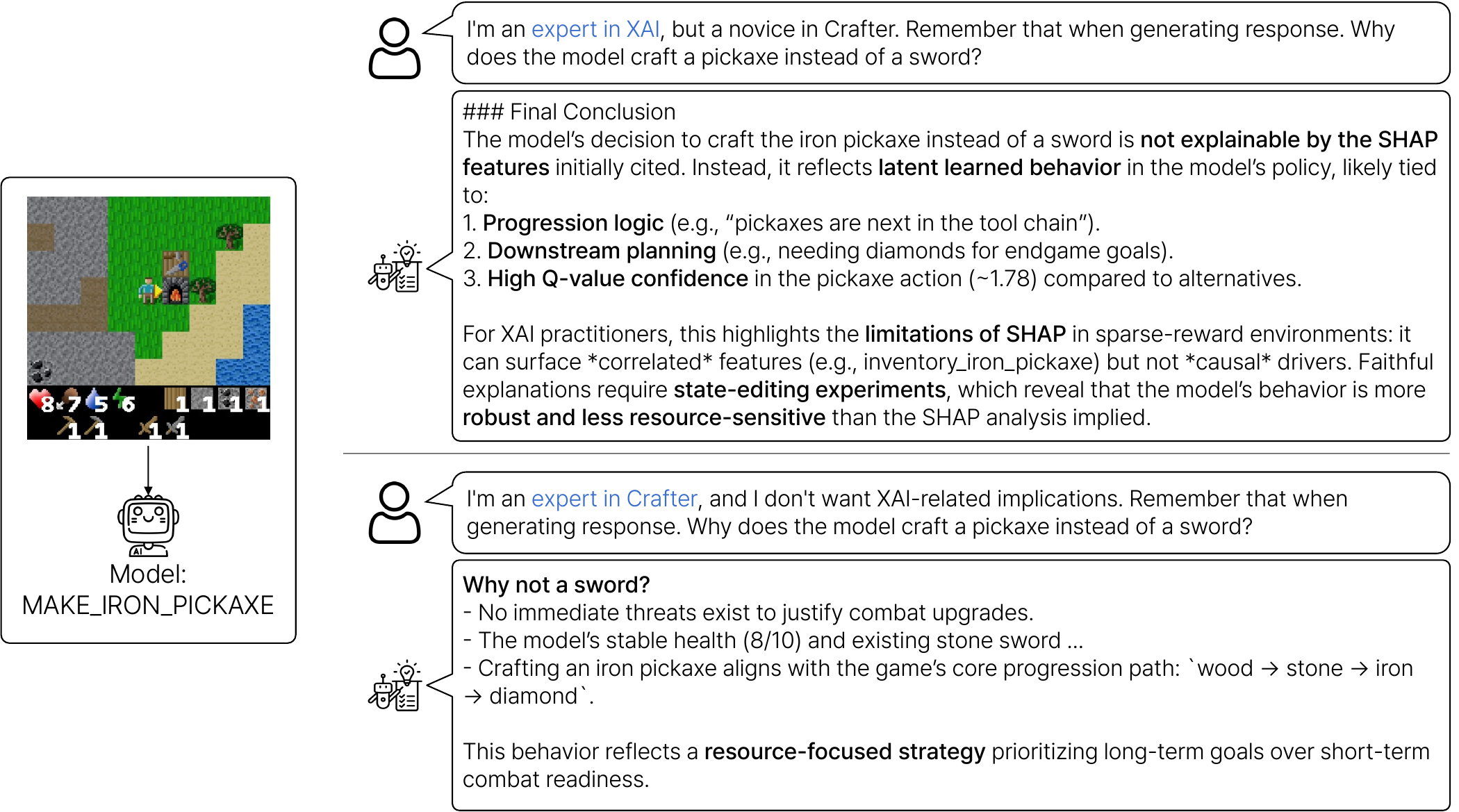}
    \caption{
        Users can specify their own background and intent in the query.
        FAX adapts the final explanation style while preserving the internal verification process.
    }
    \label{fig:analysis_user_specification}
\end{figure*}

Figure~\ref{fig:analysis_user_specification} illustrates how user expertise and intent can be incorporated into a Crafter query.
In the first case, FAX generates an explanation for an XAI expert and includes implications such as limitations of the XAI methods.
In the second case, the response omits reasoning and verification details from the XAI tools because the user request calls for a simpler answer, even though the system still verifies the relevant claims internally.
This scenario is complementary to the main faithfulness evaluation: it shows that the same benchmark environment can also test whether an agentic XAI system adapts explanation style to user needs.

\section{System prompts for Evaluation}
\label{appendix:prompt_eval}
Figures~\ref{fig:appendix_system_prompt_faith}, \ref{fig:appendix_system_prompt_info}, \ref{fig:appendix_system_prompt_relevance}, and \ref{fig:appendix_system_prompt_fluency} present the system prompts used for evaluation metrics.
These prompts follow the G-Eval-style scalar interpretation format, but are rewritten for our target criteria rather than copied from a generic G-Eval template.

\begin{figure*}[h]
    \centering
    \begin{tcolorbox}[title={\textbf{\small Implementation prompt}}, boxrule=2pt, arc=0mm]\begin{minted}[fontsize=\scriptsize, breaklines, breakanywhere, frame=lines, framesep=2mm, tabsize=4, style=vs, autogobble]{text}
You are an expert in evaluating the trustworthiness of AI model explanations.
Your task is to analyze an answer provided by an agent about a game model's behavior and generate 5 verifiable hypotheses from it.

**Context:**
- Initial State: {initial_state_desc}
- User Question: {question}
- Agent's Answer to Evaluate: {answer_to_evaluate}

**Instructions:**
1. Carefully read the agent's answer and identify the core claims or assumptions it makes about the model's behavior. (e.g., "The model attacks zombies because its health is high," or "The model avoids water because it has no boat.")
2. For each claim, devise a "what-if" scenario that can be tested using a state edit.
3. Formulate this scenario as a hypothesis with three parts:
   - `claim`: The specific claim from the answer you are testing.
   - `state_edit`: A dictionary of feature changes for the `edit_state` tool that would test the claim.
   - `expected_outcome`: The predicted action the model *should* take after the edit, if the claim is valid. The outcome should be one of the valid action names.

**Output Format:**
Provide your response as a valid JSON list of 5 dictionary objects. Do not include any text outside the JSON.

Example:
[
  {
    "state_edit": {"map(left2,up3)": "grass", "inventory_wood": 6},
    "expected_outcome": "LEFT",
  },
  ...
]
Available feature names and values for State Editing:
...
Available actions:
"NOOP", "LEFT", ...

Your JSON output:

\end{minted}
\end{tcolorbox}
        \caption{
            Evaluation prompt for Faithfulness.
            For readability, some parts are omitted and replaced with ``...''
        }
        \label{fig:appendix_system_prompt_faith}
\end{figure*}

\begin{figure*}[h]
    \centering
    \begin{tcolorbox}[title={\textbf{\small Implementation prompt}}, boxrule=2pt, arc=0mm]\begin{minted}[fontsize=\scriptsize, breaklines, breakanywhere, frame=lines, framesep=2mm, tabsize=4, style=vs, autogobble]{text}
You are a meticulous and impartial AI assistant. For this task, you must put yourself in the shoes of a human user who is trying to learn and understand the general strategy of an AI agent.

**1. Context**
The response you are evaluating is generated by an AI "Curator" that explains the behavior of a Reinforcement Learning (RL) agent in the game "Crafter". A user asks a question to understand the agent's behavior.

**2. Evaluation Goal**
Your single objective is to evaluate **Informativeness**. This means you must assess how the explanation provide information which can be used in different states.
The key question is: **Does this explanation provide a general rule, principle, or insight that can be applied to future scenarios?**
For example "The agent's next plan is mining stone." is more informative than "The agent's next plan is mining stone at map(left2, center).",
and "The agent's next plan is mining stone, and crafting a stone pickaxe." is more informative than "The agent's next plan is mining stone.".
Your evaluation is from a user's perspective. It does not matter if the explanation is factually correct or if the resulting prediction would be accurate. You are only judging how confident and able a user would feel in making a future prediction after reading the explanation.

**3. Evaluation Steps**
1. **Understand the User's Goal:** Read the `User Query` and `Final Response`. Acknowledge that the user wants to learn the agent's general strategy, not just understand a single event.
2. **Analyze the Explanation's Nature:** Analyze the content of the response. Does it describe a specific, one-time action (e.g., "The agent moved left to get the wood"), or does it reveal a broader, reusable principle (e.g., "The agent's policy is to prioritize collecting wood whenever it is nearby")?
3. **Simulate Future Prediction:** Imagine you are now shown a completely new game state. Based *only* on the explanation provided, how effectively could you form a hypothesis about the agent's next action? Does the explanation give you a "mental model" to work with?
4. **Assign a Score:** Based on this perceived predictive power and generalizability, assign a single integer score from 1 to 5 using the rubric below.

**4. Predictability Gain Rubric**
* **5 (Excellent Predictive Power):** The response provides a clear, generalizable principle or rule about the agent's behavior. A user would feel very confident applying this rule to predict actions in many new and different situations.
* **4 (Good Predictive Power):** The response provides a useful insight or pattern that could be applied to similar future situations. A user would feel reasonably confident in making predictions.
* **3 (Some Predictive Power):** The response hints at a general strategy but does not state it clearly, requiring the user to interpret heavily. It offers more than a simple description but is not a clear, actionable rule.
* **2 (Minimal Predictive Power):** The response only explains the current action in a way that is highly specific to the current state. It offers little to no insight that could be generalized to other situations (e.g., "It attacked the skeleton because it was there.").
* **1 (No Predictive Power):** The response is confusing, irrelevant, or simply describes the environment without providing any reasoning. It gives the user no basis for predicting any future actions.

**5. Input and Output Instruction**
You will be provided with a `User Query` and a `Final Response`. Your output MUST be a single integer from 1 to 5 and nothing else. Do not provide any reasoning, explanation, or additional text.
**Your final output must be only one character: "1", "2", "3", "4", or "5".**

\end{minted}
\end{tcolorbox}
        \caption{
            Evaluation prompt for Informativeness.
        }
        \label{fig:appendix_system_prompt_info}
\end{figure*}

\begin{figure*}
    \centering
    \begin{tcolorbox}[title={\textbf{\small Implementation prompt}}, boxrule=2pt, arc=0mm]\begin{minted}[fontsize=\scriptsize, breaklines, breakanywhere, frame=lines, framesep=2mm, tabsize=4, style=vs, autogobble]{text}
You are a meticulous and impartial AI assistant serving as an expert evaluator. Your task is to assess one specific criterion: **Query Relevance**.

**1. Context**
The response you are evaluating is generated by an AI "Curator" that explains the behavior of a Reinforcement Learning (RL) agent in the game "Crafter". Users ask questions about the agent's decisions, and the Curator provides an explanation.

**2. Evaluation Goal**
Your single objective is to determine how well the `Generated Response` directly answers the `User Query`. You will assign a score from 1 to 5 based *only* on the relevance rubric below.

**3. Evaluation Steps**
1. Read the `User Query` to understand the user's exact intent.
2. Read the `Generated Response`.
3. Compare the response directly against the query to judge its relevance.
4. Choose a single integer score from 1 to 5 that best represents the relevance.

**4. Query Relevance Rubric**
* **5:** The response directly and completely answers the user's question without any unnecessary information.
* **4:** The response accurately answers the user's question but may contain minor irrelevant details.
* **3:** The response addresses only a part of the user's question or provides an incomplete answer.
* **2:** The response is on the same general topic as the query but fails to answer the core question.
* **1:** The response completely ignores the user's question and is unrelated.

**5. Output Instruction**
You will be provided with a `User Query` and a `Generated Response`. Your output MUST be a single integer from 1 to 5 and nothing else. Do not provide any reasoning, explanation, or additional text.
**Your final output must be only one character: "1", "2", "3", "4", or "5".**

\end{minted}
\end{tcolorbox}
    
    \caption{Evaluation prompt for Query relevance.}
    \label{fig:appendix_system_prompt_relevance}
\end{figure*}

\begin{figure*}
    \centering
    \begin{tcolorbox}[title={\textbf{\small Implementation prompt}}, boxrule=2pt, arc=0mm]\begin{minted}[fontsize=\scriptsize, breaklines, breakanywhere, frame=lines, framesep=2mm, tabsize=4, style=vs, autogobble]{text}
You are a meticulous and impartial AI assistant serving as an expert evaluator. Your task is to assess one specific criterion: **Fluency**.

**1. Context**
The response you are evaluating is generated by an AI "Curator" that explains the behavior of a Reinforcement Learning (RL) agent in the game "Crafter".

**2. Evaluation Goal**
Your single objective is to evaluate the linguistic quality of the `Generated Response`. You will assess its grammar, structure, and clarity, assigning a score from 1 to 5 based *only* on the fluency rubric below. **Crucially, the response must be in natural, human-readable language. Responses consisting of raw data, code, or unformatted lists should be heavily penalized.** The relevance of the response to any query should be ignored.

**3. Evaluation Steps**
1. Read the `Generated Response` carefully.
2. Analyze its grammatical correctness, clarity, and overall readability.
3. Determine if the response is presented as natural language.
4. Choose a single integer score from 1 to 5 that best represents its linguistic fluency based on the rubric.

**4. Fluency Rubric**
* **5:** The response is perfectly written. It is grammatically correct, well-structured, clear, and uses natural language.
* **4:** The response is well-written and easy to understand, with only very minor errors that do not impact readability.
* **3:** The response is generally understandable but has noticeable grammatical errors or awkward phrasing.
* **2:** The response is difficult to read due to significant grammatical errors or unnatural language. **This score should also be used if the response is not primarily natural language (e.g., a raw list of keywords, unformatted data).**
* **1:** The response is grammatically incorrect, nonsensical, or unreadable. **This score must be used if the response consists entirely of non-natural language content like a code block, a JSON object, or a stack trace.**

**5. Output Instruction**
You will be provided with a `User Query` and a `Generated Response`. You must evaluate the fluency of the response only. Your output MUST be a single integer from 1 to 5 and nothing else. Do not provide any reasoning, explanation, or additional text.
**Your final output must be only one character: "1", "2", "3", "4", or "5".**

\end{minted}
\end{tcolorbox}
    \caption{Evaluation prompt for Fluency.}
    \label{fig:appendix_system_prompt_fluency}
\end{figure*}

\section{Code Release and Licensing}
\label{appendix:code_release_license}

Our code is available at \codeurl.
The release contains the implementation of FAX and all baselines, the CRAFTER-XAI-Bench tool server, scenarios, and evaluation, the tabular experiments, the three trained policy checkpoints, and all generated explanations and scores reported in this paper, together with scripts that recompute every table from these outputs.
The policy training code is not included.
Crafter is released under the MIT License.
Because CRAFTER-XAI-Bench is built on Crafter, we release our implementation code under the MIT License as well.
Redistributed Crafter code or assets retain the original Crafter copyright and MIT permission notice, and other third-party components remain governed by their respective licenses.

\section{Disclaimer about LLM usage in paper writing}

We used LLM-based assistants to polish the text and to help organize the released code.
We did not use them for other purposes, including research ideation or paper discovery.

\end{document}